 \documentclass[final,1p,times]{elsarticle}

\usepackage[T1]{fontenc}
\usepackage{lmodern}         

\usepackage{amsmath, amssymb, amsfonts}
\usepackage{amsthm}          
\usepackage{url}       
\usepackage{hyperref}       
\theoremstyle{plain}
\newtheorem{theorem}{Theorem}
\newtheorem{proposition}{Proposition}

\theoremstyle{remark}
\newtheorem{example}{Example}
\newtheorem{remark}{Remark}
\newtheorem{lemma}{Lemma}

\theoremstyle{definition}
\newtheorem{definition}{Definition}

\usepackage{enumitem}

\usepackage{graphicx}
\usepackage{multirow}
\usepackage{multicol}  
\usepackage{booktabs}

\usepackage{algorithm}
\usepackage{algorithmicx}
\usepackage{algpseudocode}

\usepackage{listings}

\usepackage{xcolor}
\usepackage{comment}

\usepackage{mathrsfs}

\usepackage[title]{appendix}

\newcommand{\KK}{\mathbb{K}}

\newcommand{\HH}{\mathbb{H}}

\begin{document}

\begin{frontmatter}



\title{A construction of an optimal base for 
conditional attribute and attributional condition implications in triadic contexts.
} 




\author[1]{Romuald Kwessy Mouona}
\ead{romualdkwessy@gmail.com}
\affiliation[1]{organization={Department of Mathematics, Faculty of Science, University of Yaounde 1}, 
            city={Yaounde}, 
            country={Cameroon}}

\author[2]{Blaise Blériot Koguep Njionou}
\ead{blaise.koguep@univ-dschang.org}
\affiliation[2]{organization={Department of Mathematics and Computer Science, Faculty of Science, University of Dschang}, 
           city={Dschang},
            country={Cameroon}}

\author[3]{Etienne Romuald Temgoua Alomo}
\ead{retemgoua@gmail.com}
\affiliation[3]{organization={Department of Mathematics, Higher Teacher Training College, University of Yaounde 1}, 
            city={Yaounde}, 
            country={Cameroon}}

\author[5]{Rokia Missaoui}
\ead{rokia.missaoui@uqo.ca}
\affiliation[5]{organization={Department of Computer Science and Engineering, Université du Québec en Outaouais (UQO)},
            addressline={101, rue Saint‐Jean‐Bosco},
             city={Gatineau (Québec)}, 
            postcode={J8X 3X7}, 
            country={Canada}}
\author[4]{Leonard Kwuida\corref{cor4}}
\ead{leonard.kwuida@bfh.ch}
\cortext[cor4]{Corresponding author}
\affiliation[4]{organization={School of Business, Bern University of Applied Sciences}, 
            addressline={Brückenstrasse 73}, 
            city={Bern}, 
            postcode={3005}, 
            country={Switzerland}}
            
\begin{abstract}
 This article studies implications in triadic contexts. Specifically, we focus on those introduced by Ganter and Obiedkov, namely conditional attribute and attributional condition implications. Our aim is to construct an optimal base for these implications.
\end{abstract}

\begin{highlights}
\item We construct an optimal set of implications for triadic contexts, by augmentation.
\item We analyze the complexity of our construction's method.
\end{highlights}

\begin{keyword}
Triadic context \sep triadic implication bases \sep pseudo-intent\sep quasi-feature \sep pseudo-feature \sep simplification logic.


\MSC[2020]  06A15 \sep 68T30 \sep 03G10.
\end{keyword}

\end{frontmatter}



\section{Introduction}\label{intro}
 A formal context is a triple $(G, M, I)$ formed by two sets $G$ (of objects) and $M$ (of attributes), and a binary relation $I$ between them, i.e. $I\subseteq G\times M$. In formal contexts, attribute implications are used to extract information about the dependencies between attributes. Thus, an implication is a relation between two sets of attributes $A$ and $B$, denoted by $A \rightarrow B$, and is \textbf{valid} if, whenever an object has all attributes in $A$, then it also has all attributes in $B$. Implications have been the subject of several studies \cite{Armstrong1974, DuquenneGuigues1986}, notably those of Duquenne and Guigues \cite{DuquenneGuigues1986}, which, for a given formal context, led to the construction of the canonical base of implications.
By incorporating the condition for which an object has an attribute, the notion of a formal context is extended \cite{Everaert-Desmedt2011}. This has led to the development of Triadic Concept Analysis (\textbf{TCA}) as an extension of Formal Concept Analysis (\textbf{FCA}) \cite{GanterObiedkov2004, Wille1995}. A triadic context is defined as a quadruple $\mathbb{K}:=(G,M, \mathcal{C}, I)$, where $G$ is a set of objects, $M$ is a set of attributes, $\mathcal{C}$ is a set of conditions, and $I$ is a relation between objects, attributes, and conditions ($I \subseteq G{\times}M{\times}\mathcal{C}$). In this article, we focus on implications of triadic contexts, which are specific connections between subsets of $M$ and $\mathcal{C}$ \cite{Biedermann1997, GanterObiedkov2004, MissaouiKwuida2011, Missaoui2020, Missaoui2022, Ruas2021}; they were introduced in the triadic framework by Biedermann \cite{Biedermann1997}.  Ganter and Obiedkov \cite{GanterObiedkov2004} extended this work by defining other types of implications. Implications in triadic contexts fall into two categories, namely Biedermann, and Ganter \& Obiedkov ones. 

Those defined by Biedermann are the following:
\begin{description}
    \item[$\star$] \emph{Biedermann’s conditional attributes implications}  (or BCAI for short), denoted by $(A_1 \rightarrow A_2)_C$, where $A_1\subseteq M$ is the premise, $A_2 \subseteq M$ is the conclusion, and $C \subseteq \mathcal{C}$ is the set of conditions (constraint). They can be interpreted as 'if an object of $G$ has all attributes in $A_1$ under all conditions in $C$, then it also has all attributes in $A_2$ under the same conditions'. They are seen as knowledge from the point of view of attributes.
    \item[$\star$] \emph{Biedermann’s attributional conditions implications} (or BACI for short), denoted by $(C_1 \rightarrow C_2)_A$ , where $C_1\subseteq \mathcal{C}$ is the premise, $C_2 \subseteq \mathcal{C}$ is the conclusion and $A \subseteq M$ is the constraint. They can be interpreted as 'if an object of $G$ has all attributes in $A$ under all conditions in $C_1$, then it also has all attributes in $A$ under all conditions in $C_2$'. 
    They are seen as knowledge from the point of view of conditions.
\end{description}
 
 The ones defined by Ganter and Obiedkov are:
\begin{description}
   \item[$\star$]  \emph{Attribute}$\times$\emph{condition implications} (or A$\times$CI for short), are of the form $E\rightarrow F$, where $E$ (premise) and $F$ (conclusion) are subsets of $M{\times}\mathcal{C}$, and interpreted as: “any object $g \in G$ in relation with all attribute-condition pairs in $E$ is also in relation with all attribute-condition pairs in $F$”.
  \item[$\star$] \emph{Conditional attribute implications} (CAI for short), denoted by $A_{1}\overset{C}{\rightarrow}A_{2}$, with $A_{1}, A_{2} \subseteq M$ and $C\subseteq \mathcal{C}$, are interpreted as: “if an object $g \in G$ has all attributes in $A_{1}$ under all set of conditions $X \subseteq C$, then $g$ also has all attributes in $A_{2}$ under $X$”. 
   \item[$\star$] \emph{Attributional condition implications} (ACI for short), denoted by $C_{1} \overset{A}{\rightarrow}C_{2}$, with $C_{1}, C_{2} \subseteq \mathcal{C}$ and $A \subseteq M$, are interpreted as: “whenever an object $g \in G$ has under the conditions in $C_{1}$ all attributes in $X \subseteq A$, then $g$ also has under the conditions in $C_{2}$ all attributes in $X$”, for all $X\subseteq A$.
\end{description} 

This article is an extension of the work carried out in \cite{RomualdBlaiseLeonardEtienne2025}. It focuses on CAI and ACI because they are more compact and convey richer semantics than BACI and BCAI. Here, we present three key notions: \textbf{feature, quasi-feature and pseudo-feature} and we show that pseudo-features correspond to the smallest family likely to generate a minimal and optimal basis of BCAI and BACI \cite{RomualdBlaiseLeonardEtienne2025}. We then introduce the notion of \textbf{unit pseudo-feature} and show that it corresponds to the smallest family of elements generating a minimal and optimal basis of CAI and ACI. Finally, we propose an algorithm for performing these constructions, followed by a theoretical study of its complexity.

The rest of this document is organized as follows. Section~\ref{basic} introduces some basic notions on FCA and TCA. 
In Section~\ref{augmentation}, we show how triadic context augmentation can contribute to the construction of quasi-features. Then we construct in Section~\ref{quasi-features and implications} complete bases and minimal bases of BCAI, BACI, CAI and ACI respectively, using quasi-features, pseudo-features and unit pseudo-feature, and provide a construction of their optimal bases.
Section~\ref{algorithm description} provides an algorithm for constructing those bases and studies its complexity. The paper ends with a conclusion.

\section{Basic notions} \label{basic}
FCA was introduced by Wille in \cite{Wille1982}, based on the understanding that a concept is constituted by its extent and intent. 
Indeed, to formalize the notion of concept, a universe of discourse or \textbf{dyadic formal context} is set by a triple $(G, M, I)$ consisting of two sets $G$ (of objects) and $M$ (of attributes) and a binary relation $I\subseteq G\times M$. A \textbf{concept} of $(G,M,I)$ is a pair $(A,B)$ such that $A \subseteq G$, $B \subseteq M$, $A'=B$ and $B'=A$, where $A'$ (the set of all attributes common to all objects in $A$) and $B'$ (the set of all objects sharing all attributes in $B$) are computed using the derivation operator $'$ defined as follows: \\
$$A' :=\{m \in M; (a,m) \in I, \ \forall a \in A\} \text{ and } B':=\{g \in G; (g,b) \in I, \ \forall b \in B\}.$$
For a dyadic concept $(A,B)$, the set $A$ is called the extent and $B$ the intent of $(A,B)$. The set of all concepts is ordered by the relation: 
$${(A_1,B_1) \leq (A_2,B_2) :\iff  A_1 \subseteq A_2\quad (:\iff  B_2 \subseteq B_1)}$$
and forms a complete lattice called the \textbf{concept lattice} of $(G, M, I)$, and denoted by $\underline{\mathfrak{B}}(G,M,I)$.
\begin{example}\label{example1}
    The table below represents a context in which the objects are clients, $G = \{1, 2, 3, 4, 5\}$, the attributes are products, $M = \{a, b, c, d\}$, where $a=accessories$, $b=books$, $c=computers$ and $d=digital\,  cameras$. The relation $I$ is defined by, $(x, y) \in I$ if and only if, the client $x$ orders the product $y$. One can verify that 
    $(\{1,3,4\},\{a,b,d\})$ is a concept of this context.
\begin{figure}[h!]
\begin{center}
\begin{tabular}{cc}
\begin{minipage}{.3\linewidth}
  \begin{center} 
\begin{tabular}{|c||c|c|c|c|}
\hline
   & a  &  b &  c  &  d  \\
\hline
\hline
1 & $\times$ &  $\times$ &  &  $\times$   \\
\hline
2 & $\times$ &   &   &  $\times$  \\
\hline
3 & $\times$ & $\times$ &  &  $\times$    \\
\hline
4 & $\times$ & $\times$ &  &  $\times$    \\
\hline
 5 & $\times$ &  &  &  $\times$    \\
\hline
\end {tabular}
\end{center}
\end{minipage} 
    &
\begin{minipage}{.4\linewidth}
  \begin{center} 
\begin{tabular}{|c||r|}
\hline
1 & abd   \\
\hline
2 & ad  \\
\hline
3 & abd   \\
\hline
4 & abd  \\
\hline
5 & ad   \\
\hline
\end {tabular}
\end{center}
\end{minipage} 
\end{tabular}
 \caption{Left: a dyadic formal context ; Right: a simplified representation}\label{formal context} 
 \end{center}
 \end{figure}
 \vspace{-0.75cm}
 
\end{example}

From Example~\ref{example1}, we can add a third dimension named \textit{Suppliers} and study the relation between Clients, Products, and Suppliers. This representation has motivated Lehmann and Wille \cite{LehmannWille1995} in 1995 to extend FCA to TCA. 

A \textbf{triadic context} is a triple denoted
$\mathbb{K}:=(G, M, \mathcal{C}, I)$, where $I\subseteq G{\times}M{\times}\mathcal{C}$ is a relation between objects in $G$, and the attributes in $M$ under conditions in $\mathcal{C}$. 

The conditions are understood in \cite{LehmannWille1995} as valuations, modalities, meanings, purposes, and reasons concerning connections between objects and attributes. 
A triadic context can be represented by a table. The two tables in Fig.~\ref{tab 1} actually represent the same example. 
taken from \cite{MissaouiKwuida2011} as an adaptation of a table in \cite{GanterObiedkov2004}. 
\begin{example}\label{example2} 
If the suppliers are Peter, Nelson, Rick, Kevin, and Simon, we can set $\mathcal{C} := \{P,N,R,K, S\}$ as our set of conditions. 
Note that the elements of $\mathcal{C}$ are capital letters, as the initials of proper nouns. The relation $I$ is then defined by $(x, y, z) \in I$ if and only if the client $x$ orders the product $y$ from the supplier $z$. The value\footnote{We will quite often use simplified notations for sets by omitting the brackets and commas. For example, we write $abc$ for $\{a, b, c\}$ or $d{\times}PN$ for $\{d\}{\times}\{P,N\}$.} $ac$ in row $1$ and column $R$ in the table in Figure~\ref{tab 1} (left) means that the client $1$ ordered the products $a$ and $c$ from the supplier $R$. In the table shown in Fig.~\ref{tab 1} (right), $PNRKS$ in Line $1$ and Column $a$ means that the client $1$ ordered the product $a$ from all suppliers.

\begin{figure}[h!]
\begin{center}
\includegraphics[scale=0.4]{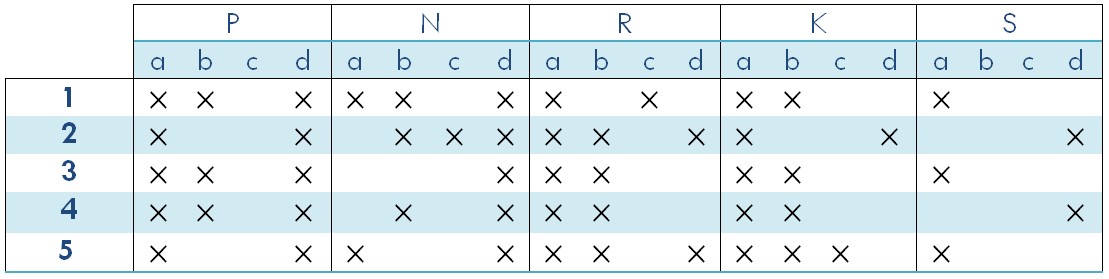}
\caption{A triadic context}\label{tab running context}
 \end{center}
 \end{figure}
 \vspace{-1cm}

\begin{figure}[h!]
\begin{center}
\begin{tabular}{cc}
\begin{minipage}{.4\linewidth}
\begin{tabular}{|c||c|c|c|c|c|}\hline
   & $P$ &  $N$ &  $R$   &   $K$  &  $S$      \\ \hline \hline
$1$ & $abd$ &  $abd$ &  $ac$   &   $ab$  &  $a$  \\ \hline
$2$ & $ad$ &  $bcd$ &  $abd$   &   $ad$  &  $d$  \\ \hline
$3$ & $abd$ &  $d$ &  $ab$  &   $ab$  &  $a$ \\ \hline
$4$ & $abd$ &  $bd$ &  $ab$  &   $ab$ &  $d$  \\ \hline
$5$ & $ad$ &  $ad$ &  $abd$  &   $abc$  &  $a$  \\ \hline
\end {tabular}
\end{minipage} 
    &
\begin{minipage}{.4\linewidth}
\begin{tabular}{|c||c|c|c|c|} \hline
   & $a$ &  $b$ &  $c$   &  $d$       \\ \hline\hline
$1$ &  $PNRKS$ &  $PNK$ &  $R$  &  $PN$      \\ \hline
$2$ & $RPK$ &  $NR$ &  $N$   &   $PNRKS$     \\ \hline
$3$ & $PRKS$ &  $RPK$ &     &  $PN$          \\ \hline
$4$ & $PRK$ &  $PNRK$ &     &   $PNS$        \\ \hline
$5$ & $PNRKS$ &  $KR$ &  $K$  &   $PNR$  \\ \hline
\end{tabular}
\end{minipage} 
\end{tabular}
 \caption{Representations of the triadic context of Fig.~\ref{tab running context}; Left context: Object-condition simplified and right context: object-attribute simplified}\label{tab 1} 
 \end{center}
 \end{figure}
 \vspace{-0.5cm}
\end{example}


 From a triadic context $\KK:=(G, M, \mathcal{C}, I)$  we can extract the following dyadic contexts: $\KK^{(1)}:=(G, M{\times}\mathcal{C}, I^{(1)})$, $\KK^{(2)}:=(M, G{\times}\mathcal{C}, I^{(2)})$, and $\KK^{(3)}:=(\mathcal{C}, G{\times}M, I^{(3)})$, where
    $$(o, (a, c)) \in I^{(1)} \iff (a, (o, c)) \in I^{(2)} \iff (c, (o, a)) \in I^{(3)} \iff (o, a, c) \in I.$$ 
     Their derivations are called \textbf{$i$-derivation}, $i \in \{1,2,3\}$.
    Additionally, we can extract the following dyadic contexts:  $(G, M, I_{C}^{12})$ with $C \subseteq \mathcal{C}$, $(G, \mathcal{C}, I_{A}^{13})$ with $A \subseteq M$  and $(M, \mathcal{C}, I_{O}^{23})$ with $O \subseteq G$, where
 \begin{eqnarray*}
     (o, a) \in I_{C}^{12} :\iff (o, a, c) \in I, \ \ \text{for all} \quad c \in C;  \\
     (o, c) \in I_{A}^{13} :\iff (o, a, c) \in I, \ \ \text{for all} \quad a \in A;  \\
     (a, c) \in I_{O}^{23} :\iff (o, a, c) \in I, \ \ \text{for all} \quad o \in O.
 \end{eqnarray*}

 Their derivations are: \textbf{$(1, 2, C)$-derivation} for $I_{C}^{12}$ with $C\subseteq \mathcal{C}$, \textbf{$(1, 3, A)$-derivation} for $I_{A}^{13}$ with $A\subseteq M$, \textbf{$(2, 3, O)$-derivation} for $I_{O}^{23}$ with $O\subseteq G$. For example, if $O\subseteq G$, 
 $$O^{(1)}=\{(a,c)\in M{\times}\mathcal{C}; (o,a,c) \in I, \  \forall o \in O \}\quad 
 \text{ and } $$    $$
 O^{(1,2,C)} =\{a \in M; (o,a,c) \in I, \  \forall (o,c) \in O{\times}C \}.$$ 

For any $X, Y$ subsets of $G$,  $M$ or  $\mathcal{C}$ and $i \in \{1,2,3\}$, we have $X\subseteq X^{(i)(i)}$, and if  $X\subseteq Y$, then $Y^{(i)} \subseteq X^{(i)}$.   
If $X_1\subseteq Y_1$ and $X_2 \subseteq Y_2$, then $(Y_1{\times}Y_2)^{(i)} \subseteq (X_1{\times}X_2)^{(i)}$.
These relations are valid if we replace the i-derivation by the $(1, 2, C)$-, $(1, 3, A)$- or $(2, 3, O)$-derivation.

A \textbf{triadic concept} of $\KK$ \cite{Ananias2021, CerfBessonRobardetBoulicaut2008} is a 3-tuple $(O, A, C) \in 2^{G}{\times}2^{M}{\times}2^{\mathcal{C}}$ 
such that $O = (A{\times}C)^{(1)}$, $A = (O{\times}C)^{(2)}$, $C = (O{\times}A)^{(3)}$.
We call $O$, $A$, $C$ and $A{\times}C$ respectively \textbf{extent}, \textbf{intent}, \textbf{modus} and \textbf{feature} 
of the concept $(O,A,C)$.  
\def\ext{\mathrm{ext}}
\def\int{\mathrm{int}}
\def\modus{\mathrm{modus}}
\begin{example}\label{concepts}
  In the context of Fig.~\ref{tab running context}, $2{\times}d{\times}PN \subsetneq 123456{\times}d{\times}PN \subseteq I$. Thus, $(2, d, PN)$ is not a triadic concept, but $(12345, d, PN)$ is a triadic concept because  
  \begin{align*}
		\{1,2,3,4,5\}&={(\{d\}{\times}\{P,N\})}^{(1)} \\
        \{d\}&={(\{1,2,3,4,5\}{\times}\{P,N\})}^{(2)} \\
        \{P,N\}&={(\{1,2,3,4,5\}{\times}\{d\})}^{(3)}
    \end{align*}
\end{example}
\def\feat{\mathrm{feat}}
For any concept $\mathit{c}$, we denote by $\ext(\mathit{c})$, $\int(\mathit{c})$, $\modus(\mathit{c})$, $\feat(\mathit{c})$ respectively its extent, intent, modus and feature. $\mathfrak{T}(\KK)$ denotes the set of all concepts and $\mathfrak{F}(\KK)$ the set of all features of $\KK$. Note that each feature defines a unique concept. 


We recall that in a finite dyadic context $(G, M, I)$, a set $P\subseteq M$ is a \textbf{pseudo-intent} if it is not closed, but contains the closure of any other pseudo-intent it contains. 
The set $\{P \to P'' \mid P\subseteq M$ is  pseudo-intent$\}$ forms an implication base of $(G,M,I)$, called the \textbf{stem base} \cite{Ganter2010,GanterWille1997,DuquenneGuigues1986}; it is also a base with the smallest cardinality. The recursive definition of pseudo-intent makes it computationally expensive to directly check whether a set is a pseudo-intent.
   Sebastian Rudolph provided in \cite{Rudolph2007} an optimized algorithm for the pseudo-intent verification. Indeed, he introduced the notion of incrementor and used it to provide a non-recursive characterization of pseudo-intents. To achieve this, for a given subset of attributes $P \subseteq M$, he added a new object $o_P$ such that $o_P'=P$, and therefore turned $P$ into an intent in the augmented context; this context is said to be augmented by $P$. He called \textbf{incrementor} any set of attributes that produces by augmentation just one new concept.
   He observed that any pseudo-intent is an incrementor and an incrementor $P$ is  a pseudo-intent if for every incrementor $Q \subseteq P$, there is an intent $R$ such that $Q \subseteq R \subseteq P$. 
  
In \cite{RomualdBlaiseLeonardEtienne2025} we provide an extension of this construction and a characterization for BCAI and BACI implications. However, no such construction has been proposed for CAI and ACI implications; yet, we know that the latter form allows a compact representation of implications, i.e. a representation of implications of the form $X \overset{B}{\rightarrow} Y$ (which correspond respectively to a family of implications described by $\{(X \rightarrow Y)_{\{b\}} : b \in B\}$). It is therefore important for us to propose such a construction for CAI and ACI implications.

 In addition, these constructions require the process of context augmentation, which plays several roles: 
\begin{description}
    \item[$\star$] It highlights any augmentation to the context;
    \item[$\star$] It allows the construction of pseudo-features which are the counterparts of pseudo-intents in dyadic contexts.
\end{description}

Since each augmentation adds new relations to the context, all concepts and implications of the augmented context must be re-computed. In this paper, we will focus on building implications of the augmented context from implications of the initial context.

  We start with the augmentation process for triadic contexts. 
  
\section{Augmentation of a triadic context} \label{augmentation}
Unless otherwise stated, we assume that $\mathbb{K}:=(G,M, \mathcal{C}, I)$ is a finite triadic context. 
%
 A context can be augmented by an attribute, an object, a condition or several of these elements simultaneously. However, the process and properties associated with augmentation remain the same, as exchanging the positions of objects, attributes and conditions does not change the relation of the context. In this section, we illustrate this process by considering the augmentation by a new object. It should be recognized that an augmentation by several elements (attributes, objects or conditions) corresponds to a sequence of augmentations of an element.
 
\begin{definition}
      Let $Z \subseteq M{\times}\mathcal{C}$. The \textbf{triadic context} 
      $\mathbb{K}=(G,M, \mathcal{C}, I)$
      \textbf{augmented} by $Z$ is the context $\KK[Z]:=(G\cup\{ o_Z\}, M, \mathcal{C}, I_Z)$ with $o_Z\notin G$ and $I_Z=I\cup(\{o_Z\}{\times}Z)$. 
\end{definition} 
\begin{example}
From Fig.~\ref{tab running context} we produce two augmentations with  $Z=d{\times}PN$ (Fig.~\ref{fig:two augmentations}~left) and $Z= (d{\times}PN) \cup (ac{\times}PRS)$ (Fig.~\ref{fig:two augmentations}~right).
\begin{figure}[h!] 
\begin{tabular}{c c}
   \begin{minipage}{.4\linewidth}
\begin{tabular}{|c||c|c|c|c|c|}
\hline
 $\KK[Z]$ & $\textcolor{blue}{P}$ &  $\textcolor{blue}{N}$ &  $R$ &   $K$  &  $S$     \\
\hline
\hline
$1$ & $abd$ &  $abd$ &  $ac$  &   $ab$  &  $a$  \\
\hline
$2$ & $ad$ &  $bcd$ &  $abd$  &  $ad$  &  $d$  \\
\hline
$3$ & $abd$ &  $d$ &  $ab$  &   $ab$  &  $a$  \\
\hline
$4$ & $abd$ &  $bd$ &  $ab$  &   $ab$  &  $d$  \\
\hline
$5$ & $ad$ &  $ad$ &  $abd$   &   $abc$ &  $a$  \\
\hline
\textcolor{blue}{ $ o_{Z}$} & $\textcolor{blue}{d}$ & $\textcolor{blue}{d}$ &     &     &    \\
\hline
\end{tabular}
   \end{minipage}
\hspace{1,5cm}

  \begin{minipage}{.4\linewidth}
\begin{tabular}{|c||c|c|c|c|c|}
\hline
 $\KK[Z]$  & $\textcolor{blue}{P}$ &  $\textcolor{blue}{N}$ &  $\textcolor{blue}{R}$  &  $K$  &  $\textcolor{blue}{S}$    \\
\hline
\hline
$1$ & $abd$ &  $abd$ &  $ac$  &   $ab$  &  $a$  \\
\hline
$2$ & $ad$ & $bcd$ &  $abd$  &   $ad$  &  $d$  \\
\hline
$3$ & $abd$ &  $d$ &  $ab$   &   $ab$  &  $a$ \\
\hline
$4$ & $abd$ &  $bd$ &  $ab$   &   $ab$  &  $d$  \\
\hline
$5$ & $ad$ &  $ad$ &  $abd$   &   $abc$  &  $a$  \\
\hline
\textcolor{blue}{$ o_{Z}$} & $\textcolor{blue}{acd}$ & $\textcolor{blue}{d}$ & $\textcolor{blue}{ac}$  &     & $\textcolor{blue}{ac}$  \\
\hline
\end{tabular}
\end{minipage}
\end{tabular}
\caption{Two augmentations: $Z=d{\times}PN$ (left) and $Z=(d{\times}PN) \cup (ac{\times}PRS)$ (right)} 
\label{fig:two augmentations}
\end{figure}
 \end{example}
In what follows, 
we investigate on the link between the derivation in $\KK$ and $\KK[Z]$
(Proposition~\ref{prop: i-deriv KZ})
and the link between their features (Proposition~\ref{propo2}).
We will denote by $^{(i_{_Z})}$ the i-derivation in $\KK[Z]$ to distinguish  it from the i-derivation in $\KK$.
\begin{proposition}\label{prop: i-deriv KZ}\cite{RomualdBlaiseLeonardEtienne2025}
Let $Z \subseteq M{\times}\mathcal{C}$, $O \subseteq G$ and $P \subseteq M{\times}\mathcal{C}$.
\begin{itemize}
 \item[(i)] $O^{(1_{Z})} = O^{(1)}$;
 \item[(ii)] $P^{(1)} = P^{(1_{Z})} \setminus \{o_{Z}\}$;
 \item[(iii)] $o_{Z} \notin P^{(1_{Z})} \implies P^{(1_{Z})(1_{Z})} = P^{(1)(1)}$;
 \item[(iv)] $o_{Z} \in P^{(1_{Z})} \implies P^{(1_{Z})(1_{Z})} = o_{Z}^{(1_{Z})} \cap P^{(1)(1)}=Z \cap P^{(1)(1)}$.
\end{itemize}
\end{proposition}
\begin{definition}\cite{RomualdBlaiseLeonardEtienne2025}
 $\HH:=(S_{1}, S_{2}, S_{3}, \gamma)$ is a \textbf{sub-context} of a context $\mathbb{K}=(G,M, \mathcal{C}, I)$ if $S_1 \subseteq G$, $S_2 \subseteq M$, $S_3 \subseteq \mathcal{C}$ and $\gamma = I \cap(S_1{\times}S_2\times S_3) $.  If in addition, the projection $(A_{1} \cap S_{1}, A_{2} \cap S_{2}, A_{3} \cap S_{3})$ of any concept $(A_{1}, A_{2}, A_{3})$ of $\mathbb{K}$ is a concept of $\HH$, we call $\HH$ a \textbf{compatible sub-context} of $\mathbb{K}$.
\end{definition}
\begin{example} $\KK$ is a sub-context of $\KK$[Z] for any $Z \subseteq M{\times}\mathcal{C}$. The context in Fig.~\ref{tab 1} (left) is  a compatible sub-context of the context in Fig.~\ref{fig:two augmentations} (right). 
\end{example}
 
\begin{proposition}\label{propo2}\cite{RomualdBlaiseLeonardEtienne2025}
Let $Z=A_{2}{\times}A_{3}$ with $A_{2} \subseteq M$, $A_{3} \subseteq \mathcal{C}$ and $\mathfrak{F}(\KK[Z])$ the set of all features of $\KK[Z]$. We have :
\begin{itemize}
 \item[(i)]  $Z \in \mathfrak{F}(\KK[Z])$ 
 \item[(ii)] $\mathfrak{F}(\KK) \subseteq \mathfrak{F}(\KK[Z])$;
 \item[(iii)]  If $Z \in \mathfrak{F}(\KK)$, then $\mathfrak{F}(\KK)=\mathfrak{F}(\KK[Z])$ and $\KK$ is a compatible sub-context of $\KK[Z]$. 
\end{itemize}
\end{proposition}
 
\begin{proof}
\begin{enumerate}
  \item[(i)] 
  $Z^{(1_{Z})}=\{o_{Z}\} \cup Z^{(1)}$, 
  $(Z^{(1_{Z})}\times A_{2})^{(3_{Z})}=A_{3}$ and $(Z^{(1_{Z})}\times A_{3})^{(2_{Z})}=A_{2}$. Therefore, $(Z^{(1_{Z})}, A_{2}, A_{3})$ is a concept, i.e. $Z=\feat(Z^{(1_{Z})}, A_{2}, A_{3})$.
  \item[(ii)] Let $C_{2}\times C_{3}\in \mathfrak{F}(\KK)$. Suppose $C_{1} \subseteq G$ and $(C_2 \times C_{3})^{(1)}=C_{1}$. We want to show that $C_{2}\times C_{3}\in \mathfrak{F}(\KK[Z])$. 

 If $Z_2$ is the projection of $Z$ on $M$ and $Z_3$ the projection of $Z$ on $\mathcal{C}$ where $\KK=(G,M,\mathcal{C},I)$, the following cases can be distinguished.
 \begin{description}
     \item[Case 1: $Z=C_{2}\times C_{3}$.] Obviously, $C_{2}\times C_{3} \in \mathfrak{F}(\KK[Z])$ from (i).
     \item[Case 2: $Z$ is a strict superset of $C_{2}\times C_{3}$.] $(C_2 \times C_{3})^{(1_Z)}=C_{1} \cup o_Z$.  \\ For \{j,k\}=\{2,3\}, 
            \begin{align*}
       ((C_{1} \cup o_Z) \times C_{j})^{(k_Z)} &= (C_{1} \cup o_Z)^{(1_Z,k_Z,C_j)} \\
                                  & =  C_{1}^{(1,k,C_j)} \cap o_Z^{(1_Z,k_Z,C_j)},  \quad  \text{w.r.t. Proposition~\ref{prop: i-deriv KZ}~(iv)}\\
                                  & = C_{k} \cap Z_{k}  \\                  
                                  & =  C_{k}, \quad  \text{since} \quad C_{k} \subseteq Z_{k}.
                  \end{align*}
Thus $(C_{1}\cup o_Z, C_{2},C_{3})$ is a concept of $\KK[Z]$, that is, $C_{2}\times C_{3}\in \mathfrak{F}(\KK[Z])$.     
     \item[Case 3: $Z$ is a strict subset of $C_{2}\times C_{3}$.] $(C_2 \times C_{3})^{(1_Z)}=C_{1}$ ; $(C_1 \times C_{2})^{(3_Z)}=C_{3}$ ; $(C_1 \times C_{3})^{(2_Z)}=C_{2}$. Thus, $C_{2}\times C_{3}\in \mathfrak{F}(\KK[Z])$. 
     \item[Case 4: none of the above cases is verified.] Here, the concept $(C_1,C_2,C_3)$ of $\KK$ remains the same in $\KK[Z]$, i.e. $C_{2}\times C_{3}\in \mathfrak{F}(\KK[Z])$. 
   \end{description}

\item[(iii)] $|\mathfrak{F}(\KK)| \leq |\mathfrak{F}(\KK[Z])|$ for any $Z$ and from $(i)$, $(ii)$ and the fact that $Z \in \mathfrak{F}(\KK)$, we have $|\mathfrak{F}(\KK[Z])| \leq |\mathfrak{F}(\KK)|$. Therefore, $|\mathfrak{T}(\KK[Z])| = |\mathfrak{T}(\KK)|$. Moreover, from (ii) and the fact that features and concepts are in one-to-one correspondence, it immediately follows that they have the same features. Since $\KK$ is a sub-context of $\KK[Z]$, we can conclude that $\KK$ is a compatible sub-context of $\KK[Z]$.

To prove that $ \mathfrak{F}(\KK[Z]) =  \mathfrak{F}(\KK)$, it is sufficient to show that $ \mathfrak{F}(\KK[Z]) \subseteq  \mathfrak{F}(\KK)$ since $\mathfrak{F}(\KK) \subseteq \mathfrak{F}(\KK[Z])$ from $(ii)$. Let $ Z_{1} \in \mathfrak{F}(\KK[Z])$, since $|\mathfrak{T}(\KK[Z])| = |\mathfrak{T}(\KK)|$ and $\mathfrak{F}(\KK) \subseteq \mathfrak{F}(\KK[Z])$, we can write $ Z_{1} = Z \in \mathfrak{F}(\KK)$ or $Z_{1} \in \mathfrak{F}(\KK)$. It is then obvious that $Z_{1} \in \mathfrak{F}(\KK)$, i.e., $\mathfrak{F}(\KK[Z]) \subseteq  \mathfrak{F}(\KK)$.
  By the definition of a sub-context,  it is obvious that $\KK$ is a sub-context of $\KK[Z]$ since $I_{Z}=I\cup (\{o_{Z}\}{\times}Z)$.
  Finally, $\KK$ is a compatible sub-context of $\KK[Z]$ because $\mathfrak{F}(\KK[Z]) = \mathfrak{F}(\KK)$. 
\end{enumerate}
\end{proof}

\begin{remark}
The computation of the features of the augmented context is done incrementally, based on the concepts of the initial context (see \cite{GodinMissaouiAlaoui1995,MakhalovaNourine2017} for more details).
\end{remark}

As the name suggests, augmentation generates new information. However, it is essential to note that not all augmentations generate the same amount of information. 
As Sebastian Rudolf shows in \cite{Rudolph2007}, incrementors can characterize pseudo-closed sets in a dyadic context. Can we expect similar results in triadic contexts?

 \begin{definition}\cite{RomualdBlaiseLeonardEtienne2025}
   A  set $P = A_{2}{\times}A_{3} \subseteq M{\times}\mathcal{C}$ is called \textbf{quasi-feature} of $\KK$, if it is not a feature of $\KK$ and the context $\KK[P]$ contains only one new feature with respect to the concepts of $\KK$.
 \end{definition}

Observe that, if $P$ is a quasi-feature of $\KK$, then $|\mathfrak{F}(\KK[P])| = 1 + |\mathfrak{F}(\KK)|$. 
Hence, $P$ is a quasi-feature of $\KK$ if and only if $  P \notin \mathfrak{F}(\KK)$ and for any $ Z \in \mathfrak{F}(\KK[P]), \ Z = P$ or $Z \in \mathfrak{F}(\KK)$.

\begin{example}\label{exple 6} We want to show here that the product $d{\times}P$ is a quasi-feature of our running context. 
\begin{figure}[h!] 
\begin{tabular}{c c}
   \begin{minipage}{.4\linewidth}
\begin{tabular}{|c||c|c|c|c|c|}
\hline
 $\KK$ & $P$ &  $N$ &  $R$ &   $K$  &  $S$     \\
\hline
\hline
$1$ & $abd$ &  $abd$ &  $ac$  &   $ab$  &  $a$  \\
\hline
$2$ & $ad$ &  $bcd$ &  $abd$  &  $ad$  &  $d$  \\
\hline
$3$ & $abd$ &  $d$ &  $ab$  &   $ab$  &  $a$  \\
\hline
$4$ & $abd$ &  $bd$ &  $ab$  &   $ab$  &  $d$  \\
\hline
$5$ & $ad$ &  $ad$ &  $abd$   &   $abc$ &  $a$  \\
\hline
\end{tabular}
   \end{minipage}
&
  \begin{minipage}{.4\linewidth}
\begin{tabular}{|c||c|c|c|c|c|}
\hline
 $\KK[d{\times}P]$  & $\textcolor{blue}{P}$ &  $N$ &  $R$  &  $K$  &  $S$    \\
\hline
\hline
$1$ & $abd$ &  $abd$ &  $ac$  &   $ab$  &  $a$  \\
\hline
$2$ & $ad$ & $bcd$ &  $abd$  &   $ad$  &  $d$  \\
\hline
$3$ & $abd$ &  $d$ &  $ab$   &   $ab$  &  $a$ \\
\hline
$4$ & $abd$ &  $bd$ &  $ab$   &   $ab$  &  $d$  \\
\hline
$5$ & $ad$ &  $ad$ &  $abd$   &   $abc$  &  $a$  \\
\hline
\textcolor{blue}{$ o_{Z}$} & $\textcolor{blue}{d}$ &  &  &     &  \\
\hline
\end{tabular}
\end{minipage}
\end{tabular}
\caption{Our running example (left context) and its augmentation by $Z=d{\times}P$ (right context).} 
\label{example of pseudo-feature}
\end{figure}
Here, $Z=d{\times}P$ and the new object is $o_{Z}$. 
We can verify that the left context has 33 concepts\footnote{We acknowledge the use of \textit{FCA Tools Bundle} at \url{https://fca-tools-bundle.com}.}, while the right context has 34 concepts. Thus, the product $d{\times}P$ is a quasi-feature of the left context.
\end{example}

In the following, we will highlight the link between quasi-features and implications.

\section{quasi-features and implications in triadic contexts} \label{quasi-features and implications}

quasi-features are important for constructing implications. We use them here to construct bases of triadic implications of the forms BCAI, BACI, CAI and ACI. To facilitate understanding, we begin with a few reminders of these triadic implications.

\begin{definition}\cite{RomualdBlaiseLeonardEtienne2025}
    A BCAI $(A_{1} \rightarrow A_{2})_{C}$ is \textbf{valid} in $\KK$ if each time an object of $G$ has all the attributes in $A_{1}$ under all conditions in $C$, this same object also has all attributes in $A_{2}$ under the same conditions, i.e.
    \begin{eqnarray*}
        (A_{1}{\times}C)^{(1)} \subseteq (A_{2}{\times}C)^{(1)} \quad (\Longleftrightarrow A_{2} \subseteq A_{1}^{(1,2,C)(1,2,C)})
    \end{eqnarray*} 
    In a similar way, a BACI $(C_{1} \rightarrow C_{2})_{A}$ is \textbf{valid} in $\KK$ if each time an object of $G$ has all attributes in $A$ under the  conditions in $C_{1}$, then this same object also has all attributes in $A$ under all conditions in $C_{2}$, i.e.
    \begin{eqnarray*}
        (A{\times}C_{1})^{(1)} \subseteq (A{\times}C_{2})^{(1)} \quad (\Longleftrightarrow C_{2} \subseteq C_{1}^{(1,3,A)(1,3,A)})
    \end{eqnarray*} 
\end{definition}  
\begin{example}
    In our running context (Fig.~\ref{tab 1}), we have the valid implications: $(d \rightarrow a)_{P}$ and $(P \rightarrow KP)_{b}$ since $(d{\times}P)^{(1)} \subseteq (a{\times}P)^{(1)}$ and $(b{\times}P)^{(1)} \subseteq (b{\times}KP)^{(1)}$.
\end{example}      

  \begin{definition}
    A CAI $A_{1} \overset{C}{\rightarrow} A_{2}$ is \textbf{valid} in $\KK$ if each time an object of $G$ has all attributes in $A_{1}$ under all set of conditions $X \subseteq C$, then this same object also has all attributes in $A_{2}$ under $X$, i.e.  
    \begin{eqnarray*}
       (A_{1}{\times}X)^{(1)} \subseteq (A_{2}{\times}X)^{(1)} \quad \text{for all} \ X \subseteq C \quad (\Longleftrightarrow A_{1} \subseteq A_{2}^{(1,2,X)(1,2,X)}  \quad \text{for all} \ X \subseteq C).
    \end{eqnarray*} 
    Similarly, an ACI $C_{1} \overset{A}{\rightarrow} C_{2}$ is \textbf{valid} in $\KK$ if each time an object of $G$ has all conditions in $C_{1}$ under all set of attributes $X \subseteq A$, this same object also has all conditions in $C_{2}$ under the same attributes, i.e.
    \begin{eqnarray*}
      (X{\times}C_{1})^{(1)} \subseteq (X{\times}C_{2})^{(1)}  \quad \text{for all} \ X \subseteq A  \quad (\Longleftrightarrow C_{2} \subseteq C_{1}^{(1,3,X)(1,3,X)} \quad \text{for all} \ X \subseteq A).
    \end{eqnarray*}
\end{definition}
\begin{remark}
    If an implication $\sigma$ is valid in a context $\KK$, then $\KK$ is called a \textbf{model} for $\sigma$. A model of a family of implications $\Sigma$ is a context in which all implications of $\Sigma$ are valid. 
\end{remark}

\begin{example}\label{example 8}
    In the context illustrated by Fig.~\ref{tab running context}, the following implications are valid : $d \overset{P}{\rightarrow} a$ and $P \overset{abc}{\rightarrow} K$; this can be justified by the inclusions: $(d{\times}P)^{(1)} \subseteq (a{\times}P)^{(1)}$ and $(x{\times}P)^{(1)} \subseteq (x{\times}K)^{(1)}$, for all $x \in \{a,b,c\}$.     
\end{example}       
  
   We recall that $\KK \models \sigma $ means that $\sigma$ is valid in $\KK$. If $\Sigma$ is a set of implications verifying $\KK \models \sigma$, for all $\sigma$ in $\Sigma$, then we can write $\KK \models \Sigma$. Finally, $\Sigma$ semantically follows from $\sigma$ (or $\Sigma \models \sigma$ for short) if and only if $\sigma$ is valid in every context, in which all implications of $\Sigma$ are valid. \\
   
In dyadic contexts, for two distinct pseudo-closed sets $P_1$ and $P_2$ such that $P_1 \subset P_2$, there exists a closed set $F$ between them ($P_1 \subset F \subset P_2$). This makes it possible to obtain two distinct implications ($P_1\to P_1'' \subseteq F$ and $P_2 \rightarrow P_2^{''}$) from $P_1$ and $P_2$. This analogy is interesting when implementing in triadic contexts. 
 \begin{lemma}\label{lemme valid implication} 
  Let 
  $X \subseteq M$ and $Y \subseteq \mathcal{C}$. The following implications are valid in $\KK$.
$$ \text{BCAI} \quad (X \rightarrow X^{(1,2,Y)(1,2,Y)} )_{_Y} \qquad  \textbf{and} \qquad  \text{BACI} \quad (Y \rightarrow Y^{(1,3,X)(1,3,X)} )_{_X}$$ 
\end{lemma}
\begin{proof}
To prove that $(X \rightarrow X^{(1,2,Y)(1,2,Y)} )_{_Y}$ is valid, we have to show that $(X{\times}Y)^{(1)} \subseteq ( X^{(1,2,Y)(1,2,Y)}{\times}Y)^{(1)}$. Since $(X^{(1,2,Y)}; X^{(1,2,Y)(1,2,Y)}; (X^{(1,2,Y)}{\times}X^{(1,2,Y)(1,2,Y)})^{(3)})$ is a concept  
 satisfying $ Y \subseteq (X^{(1,2,Y)}{\times}X^{(1,2,Y)(1,2,Y)})^{(3)}$, we can write $(X{\times} Y)^{(1)} =  X^{(1,2,Y)} = (X^{(1,2,Y)(1,2,Y)}{\times}(X^{(1,2,Y)}{\times}X^{(1,2,Y)(1,2,Y)})^{(3)})^{(1)} \subseteq ( X^{(1,2,Y)(1,2,Y)}{\times}Y)^{(1)}$. Finally,  \\ $(X{\times}Y)^{(1)} \subseteq ( X^{(1,2,Y)(1,2,Y)}{\times}Y)^{(1)}$.
 
 The proof of the BACI is similar. 
\end{proof}
    The product $X{\times}Y$ will essentially be considered as a quasi-feature in what follows.

\begin{example}
  We have seen in Example~\ref{exple 6} that $\{d\}{\times}\{P\}$ is a quasi-feature of the context of Fig.\ref{tab 1} (left). Since $\{d\}^{(1,2,\{P\})(1,2,\{P\})}=\{a,d\}$, the BCAI: $(d \rightarrow ad)_{P}$ is valid.
\end{example}

 Next, we characterize quasi-features that generate non-trivial implications, i.e. implications that provide meaningful information; to be more precise, 
 implications whose conclusion is not a subset of the premise, or whose condition is not empty.


\begin{definition}
 A quasi-feature $X{\times}Y$ of $\KK=(G, M, \mathcal{C}, I)$ is said to be \textbf{relevant} or \textbf{informative} with respect to $M$ (respectively, $\mathcal{C}$) if $X^{(1,2,Y)(1,2,Y)} \backslash X \neq \emptyset$ and $Y \neq \emptyset$ (respectively, $Y^{(1,3,X)(1,3,X)} \backslash Y \neq \emptyset$ and $X \neq \emptyset$).
\end{definition}

In all what follows, $\mathbb{P}_{2}(\KK)$ will be the set of all relevant quasi-features of $\KK$ with respect to $M$ and $\mathbb{P}_{3}(\KK)$ those with respect to $\mathcal{C}$.

The following outlines an interesting fact about quasi-features with one empty component.
  \begin{proposition}\label{propo 100percent-support implication and optimality}\cite{RomualdBlaiseLeonardEtienne2025}
      Let $a \in M$ and $c \in \mathcal{C}$.
\begin{enumerate}
  \item If there is a concept $\mathfrak{c}_{_c}$ with non empty components such that $c \in \modus(\mathfrak{c}_{_C})$ and $\ext(\mathfrak{c}_{_c}) = G$, then \ $ \emptyset \overset{c}{\rightarrow} \int(\mathfrak{c}_{_c})$ \ is a CAI.
  \item If there is a concept $ \mathfrak{c}_{_a}$ with non empty component such that $a \in \int(\mathfrak{c}_{_a})$ and $\ext(\mathfrak{c}_{_a}) = G$, then $ \emptyset \overset{a}{\rightarrow} \modus(\mathfrak{c}_{_a})$ \ is a ACI.
\end{enumerate} 
\end{proposition}
\begin{remark}\label{remark quasi-feature and optmality}
Let $\mathbb{K}$ be a triadic context, $\mathcal{X}$ and $\mathcal{Y}$ the sets of all such A and C respectively describe in the Proposition~\ref{propo 100percent-support implication and optimality}. Subsets of $M{\times}\mathcal{C}$ having the form $\emptyset{\times}c$, $c \in \mathcal{Y}$ summarizing all implications of the form: $m \overset{c}{\rightarrow} \int(\mathfrak{c}_{_c}) \, \ \text{for all} \ m \in M$, are all as special as sets in $\mathbb{P}_{2}(\mathbb{K})$ (respectively, Subsets of $M{\times}\mathcal{C}$ having the form $a{\times}\emptyset$, $a \in \mathcal{X}$, summarizing all implications of the form: $c \overset{a}{\rightarrow} \modus(\mathfrak{c}_{_a}) \, \ \text{for all} \ c \in \mathcal{C}$, are all as special as sets in $\mathbb{P}_{3}(\mathbb{K})$). Therefore, in all what follows, we will adopt the notation $\emptyset{\times}c$ (respectively, $a{\times}\emptyset$) to name any quasi-feature satisfying $c \in \mathcal{Y}$ (respectively, $a \in \mathcal{X}$). 
\end{remark}

 In what follows, we recall the axiomatic system for triadic implications \cite{RodriguezEncisoRokiaMora2017, RodriguezCorderoRokiaMora2016}. Throughout the rest of this document, we will use the notation $\Sigma \vdash \sigma$ to mean that 'an implication $\sigma$ is a syntactic consequence of a set of implications $\Sigma$'; in other words, '$\sigma$ can be deduced from $\Sigma$'. The following principles describe this deduction.
 
    
  For $X, Y, W, Z \subseteq M$ and $C, C_1, C_2 \subseteq \mathcal{C}$ the logic for BCAI relies on two axioms: \\
 \textbf{[Non-constraint]}   \quad  $\vdash (\emptyset \rightarrow M)_{\emptyset}$          \\
 \textbf{ [Reflexivity]}          \quad  $ \vdash  (X  \rightarrow X)_{\mathcal{C}}$          \\
  And three inference rules, which are : \\
 \textbf{[Augmentation]}  \quad  $(X \rightarrow Y)_{C}   \vdash (X \cup Z \rightarrow Y \cup Z)_{C}$          \\
 \textbf{[Transitivity]}   \quad  $\{ (X \rightarrow Y)_{C} ; \  (Y \rightarrow Z)_{C} \} \vdash  (X \rightarrow Z)_{C}$   \\
  \textbf{[Conditional composition]}          \quad  $\{ (X \rightarrow Y)_{C_{1}} ; \  (Z \rightarrow W)_{C_{2}} \} \vdash   (X \cup Z \rightarrow Y \cap W)_{C_{1} \cup C_{2}}$

Since we are dealing with Biedermann's implications, it follows that [conditional decomposition]:$(X \rightarrow Y)_{C_1 \cup C_2} \vdash (X \rightarrow Y)_{C_1}$ cannot be possible. The soundness and completeness of the above logic for Biedermann's implications derive from the study of conditional attributes implicational logic made in \cite{RodriguezCorderoRokiaMora2016}. In the following, we recall other rules that derive from the above logic (see \cite{RodriguezCorderoRokiaMora2016} for details).

  \textbf{[Decomposition]}   \quad  $(X \rightarrow Y \cup Z)_{C}  \vdash  (X \rightarrow Y)_{C}$
  
  \textbf{[Pseudotransitivity]}   \quad  $\{ (X \rightarrow Y)_{C} ; \  (Y \cup Z \rightarrow W)_{C} \} \vdash  (X \cup Z \rightarrow W)_{C}$
  
  \textbf{[Additivity]}   \quad  $\{ (X \rightarrow Y)_{C} ; \  (X \rightarrow Z)_{C} \} \vdash  (X \rightarrow Y \cup Z)_{C}$
  
   \textbf{[Accumulation]}   \quad  $\{ (X \rightarrow Y \cup Z)_{C} ; \  (Z \rightarrow W)_{C} \} \vdash  (X \rightarrow Y \cup Z \cup W)_{C}$ \\

 \begin{definition} 
  With respect to the above logic, a derivation of an implication $\sigma$ from a set of implication $\Sigma$ is a sequence $\sigma_1, \dots, \sigma_n$ of implications satisfying:
  \begin{itemize}
      \item  $\sigma_n$ is just $\sigma$
      \item  for every $i=1,2,…,n$:
      \begin{itemize}
          \item every $\sigma_i$ is in $\Sigma$ (assumption) or is an axiom
          \item or $\sigma_i$ results from $\sigma_j$, $j<i$ by applying augmentation
          \item or $\sigma_i$ results from $\sigma_j$ and $\sigma_k$ ($j,k<i$) applying transitivity or conditional composition.
      \end{itemize}
  \end{itemize}
\end{definition}
 
 We are interested in finding a family that can generate (with respect to the above logic) all valid implications of a context, i.e. a family of implications capable of generating any implication valid in the same context.

\begin{definition}\cite{Belohlavek2008}
A set of implications $\mathcal{B}$ of a context $\KK$ is \textbf{complete} if for any implication $\sigma$, we have $$\sigma \ \text{is valid in}  \ \KK \quad \text{if and only if} \quad \mathcal{B} \models \sigma.$$ 
In this study, we will refer to any complete family of implications as a \textbf{base} of implications. 

\end{definition}
From \cite{RodriguezEncisoRokiaMora2017}, we know that an implication $\sigma$ semantically follows from a set of implications $\mathcal{B}$ if $\sigma$ can be derived syntactically from $\mathcal{B}$ using the axiomatic system described above. Therefore, $\mathcal{B}$ is complete in a context $\KK$ if for all $\sigma$ valid in $\KK$, we have $\mathcal{B} \vdash \sigma$. 

We are going to recall two important propositions to better understand the construction of a complete family of implications in triadic contexts  \cite{RodriguezEncisoRokiaMora2017}. First, we need to extend the notion of closure operator known in formal contexts \cite{Rudolph2007}.
\begin{proposition}\cite{RodriguezEncisoRokiaMora2017}
    Let $\Sigma$ be a set of BCAI valid in $\KK$, $A \subseteq M$ a subset of attributes, and $C \subseteq \mathcal{C}$ a subset of conditions.  The map
 \begin{eqnarray*}
 (.)_{\Sigma, C} \ : \  2^{M}   &  \to     &   2^{M}  \\
                             A     & \mapsto  & (A)_{\Sigma, C} := \{ a \in M : \Sigma \vdash (A \rightarrow a)_{C} \}  
\end{eqnarray*}
  is a closure operator.
\end{proposition}
 
This closure operator enables us to simplify the derivation process of implications as is shown below. 

\begin{proposition}\cite{RodriguezEncisoRokiaMora2017}
  Let $\Sigma$ be a set of BCAI, $A_{1}, A_{2} \subseteq M$ subsets of attributes, and $C \subseteq \mathcal{C}$ a subset of conditions. 
  The following statements are equivalent:
  \begin{eqnarray*}
      \Sigma \vdash (A_{1} \rightarrow A_{2})_{C} \qquad \text{and} \qquad  A_{2} \subseteq (A_{1})_{\Sigma, C}
  \end{eqnarray*}
\end{proposition}

We have already shown how to extract some valid implications from a triadic context (Lemma~ \ref{lemme valid implication}). It would be interesting to build a complete subfamily of these implications. 
\begin{lemma}\label{lemme complete set of implications}\cite{RomualdBlaiseLeonardEtienne2025}
The following sets of implications are complete in $\KK$.
\begin{eqnarray*}
 \mathcal{B}_{BCAI} = \{ (X \rightarrow X^{(1,2,Y)(1,2,Y)} )_{_Y} : X{\times}Y \in \mathbb{P}_{2}(\KK) \}  \\
 \mathcal{B}_{BACI} = \{ (Y \rightarrow Y^{(1,3,X)(1,3,X)} )_{_X} : X{\times}Y \in \mathbb{P}_{3}(\KK) \}  
\end{eqnarray*}
\end{lemma}

\begin{proof}
We focus on BCAI since the results for BACI follow from interchanging $M$ and $\mathcal{C}$.
%
$\mathcal{B}_{BCAI}$ is complete if for any $A_{1}, A_{2} \subseteq M$ and $C \subseteq \mathcal{C}$,  $$ \KK \ \vDash \ (A_{1} \rightarrow A_{2})_{C}   \ \Rightarrow \     \mathcal{B}_{BCAI} \ \vdash \ (A_{1} \rightarrow A_{2})_{C}  $$ that is $$ \mathcal{B}_{BCAI} \ \nvdash \ (A_{1} \rightarrow A_{2})_{C}  \ \Rightarrow \      \KK \ \nvDash \ (A_{1} \rightarrow A_{2})_{C}$$

It is proven in \cite{RodriguezEncisoRokiaMora2017} that the inference system behind $\vdash$ is sound and complete with respect to the semantics of $\vDash$, that is \textbf{every model of a set of implications (in particular) $\mathcal{B}_{BCAI}$ is a model of $(A_{1} \rightarrow A_{2})_{C}$ if $(A_{1} \rightarrow A_{2})_{C}$ can be derived syntactically from $\mathcal{B}_{BCAI}$ using the conditional attribute simplification logic axiomatic system (which is use in this work), i.e.}
$$\mathcal{B}_{BCAI} \vDash  (A_{1} \rightarrow A_{2})_{C} \quad \Longrightarrow \quad \mathcal{B}_{BCAI} \vdash  (A_{1} \rightarrow A_{2})_{C}$$
Indeed, Assuming $\mathcal{B}_{BCAI} \ \nvdash \ (A_{1} \rightarrow A_{2})_{C}$, we obtain $C \neq \emptyset$ since $\vdash (A_{1} \rightarrow A_{2})_{\emptyset}$ from [Non-constraint]. We need to show that $\mathcal{B}_{BCAI} \ \nvDash \ (A_{1} \rightarrow A_{2})_{C}$, i.e. there is a model of $\mathcal{B}_{BCAI}$ which is not a model of $(A_{1} \rightarrow A_{2})_{C}$.
  
   Let us consider the context $\KK= (G, M, \mathcal{C}, \gamma)$ where $G = \{1, 2\}$ and $\gamma$ the relation such that:
  \begin{enumerate}
    \item $(1, m, b) \in \gamma$ if and only if one of the two conditions holds:
      \begin{description}
        \item[i.] $b \in C$ and $m \in (A_{1})_{\mathcal{B}_{BCAI}, b}$ or
        \item[ii.] $b \notin C$ and $m \in (\emptyset)_{\mathcal{B}_{BCAI}, b}$
      \end{description}
    \item $(2, m, b) \in \gamma$ for all $m \in M$ and $b \in B$. 
  \end{enumerate}
  
  ($\star$) \textbf{Firstly, we will show that all implications in $\mathcal{B}_{BCAI}$ are valid in $\KK$}.  \\
  For $(A_{3}\rightarrow A_{4})_{C_{1}} \in \mathcal{B}_{BCAI}$, we have to prove that $(A_{3}, C_{1})^{(1)} \subseteq (A_{4}, C_{1})^{(1)}$ holds in $\KK$.

  If $A_3 = \emptyset$, then $A_{4} \subseteq (\emptyset)_{\mathcal{B}_{BCAI}, C_1}$ and $(A_3, C_1)^{(1)}=\{1,2\}$. Therefore, if $C_1 \nsubseteq C$, then by the definition of $\KK$, $(A_4, C_1)^{(1)}=\{1,2\}=(A_3, C_1)^{(1)}$. Else, $C_1 \subseteq C$ and $A_4 \subseteq (\emptyset)_{\mathcal{B}_{BCAI}, C_1} \subseteq (A_1)_{\mathcal{B}_{BCAI}, C_1}$. That is, $(A_4, C_1)^{(1)}=\{1,2\}$. 
  
   If $A_3 \neq \emptyset$, then we have the following cases:
   \begin{description}
     \item[Case 1.] $C_{1} \subseteq C$. If $A_{4} \nsubseteq (A_{1})_{\mathcal{B}_{BCAI}, C_{1}}$ then, $(A_{3}, C_{1})^{(1)} = \{2\} \subseteq (A_{4}, C_{1})^{(1)}$  by definition of $\KK$. 
     
    Else, if $A_{4} \subseteq (A_{1})_{\mathcal{B}_{BCAI}, C_{1}}$ then we have $(A_{4}, C_{1})^{(1)}=G$.
    It is then obvious that $(A_{3}, C_{1})^{(1)} \subseteq (A_{4}, C_{1})^{(1)}$.
     \item[Case 2.] $C_{1} \nsubseteq C$. We have  $(A_{3}, C_{1})^{(1)} = \{2\} \subseteq (A_{4}, C_{1})^{(1)}$ by definition of $\KK$.
   \end{description}
     
    ($\star\star$) \textbf{We conclude by showing that $(A_{1} \rightarrow A_{2})_{C}$ is not valid in $\KK$.} \\
     Suppose that $\KK \models (A_{1} \rightarrow A_{2})_{C}$, that is $(A_{1}, C)^{(1)} \subseteq (A_{2}, C)^{(1)}$. Since $A_{1} \subseteq (A_{1})_{\mathcal{B}_{BCAI},C}$ we have $(\{1;2\},A_{1},C) \in \gamma$, that is, $\{1;2\} \subseteq (A_{1} , C)^{(1)} \subseteq (A_{2}, C)^{(1)}$, i.e. $(1,A_{2},C) \in \gamma$. Therefore, $A_{2} \subseteq (A_{1})_{\mathcal{B}_{BCAI},C}$. This contradicts the fact that $\mathcal{B}_{BCAI} \ \nvdash \ (A_{1} \rightarrow A_{2})_{C}$. We can finally conclude that the above equivalence is true. \\

  With respect to this equivalence, if $\mathcal{B}_{BCAI} \ \nvdash \ (A_{1} \rightarrow A_{2})_{C}$, then $\mathcal{B}_{BCAI} \ \nvDash \ (A_{1} \rightarrow A_{2})_{C}$, i.e. $(A_{1} \rightarrow A_{2})_{C}$ is not valid, in any context where every implications of $\mathcal{B}_{BCAI}$ are not valid. In particular, $(A_{1} \rightarrow A_{2})_{C}$ is not valid in $\KK$. Thus, the set $\mathcal{B}_{BCAI}$ is complete.
\end{proof}


 \begin{example}\label{exple complete base}
 Here, we are looking forward to summarizing all BCAI and BACI of our running context using Lemma~ \ref{lemme complete set of implications}. We can verify that the set of all relevant quasi-features with respect to $M$ is as follows:
 \begin{center}
 $ a{\times}P \ ; \ a{\times}N \ ; \ b{\times}P \ ;\ b{\times}N \ ; \  b{\times}R \ ; \  b{\times}K \ ; \ b{\times}S \ ; \ c{\times}P \ ;\ c{\times}N \ ; \  c{\times}R \ ; \  c{\times}K \ ; \ c{\times}S \ ;$ \\
 $ c{\times}PS \ ; \ d{\times}P \ ; \ d{\times}R \ ; \ d{\times}K \ ; \ bc{\times}S \ ; \  \emptyset {\times} P \ ; \  \emptyset {\times} N \ ; \  \emptyset {\times} R \ ; \  \emptyset {\times} K \ ; \ \emptyset {\times} PN \ ; \  \emptyset {\times} PR$ \ ; \\
   $\emptyset {\times} PK \ ; \ \emptyset {\times} RK \ ; \  \emptyset {\times} RPK \ ; \ acd{\times}P \ ; \ abd{\times}RPK \ ; \ abc{\times}RPK \ ; \ ad{\times}PNRKS$ \ ; \\  
   $ab{\times}RPKS \ ; \ abc{\times}RPKS \ ; \ abd{\times}PNRKS$.
 \end{center}

Therefore,   
  \begin{align*}
     \mathcal{B}_{BCAI} & = \{(a \rightarrow ad)_{P} \ ; \ (a \rightarrow ad)_{N} \ ;  \ (b \rightarrow abd)_{P} \ ; \ (b \rightarrow bd)_{N} \ ;  \ (b \rightarrow ab)_{R} \ ; \ (b \rightarrow ab)_{K} ; \\ 
      & \phantom{==}   (b \rightarrow abcd)_{S} \ ; \ (c \rightarrow abcd)_{P} \ ; \ (c \rightarrow bcd)_{N} \ ;  \ (c \rightarrow ac)_{R} \ ; \ (c \rightarrow abc)_{K} \ ; \\
      & \phantom{==}   (c \rightarrow abcd)_{S} \ ; \ (c \rightarrow abcd)_{PS} \ ; \ (d \rightarrow ad)_{P} \ ; \ (d \rightarrow abd)_{R} \ ; \  (d \rightarrow ad)_{K} \ ; \\
      & \phantom{==}   (bc \rightarrow abcd)_{S} \ ; \ (\emptyset \rightarrow ad)_{P} \ ; \ (\emptyset \rightarrow d)_{N} \ ; \ (\emptyset \rightarrow a)_{R} \ ; \ (\emptyset \rightarrow a)_{K} \ ; \ (\emptyset \rightarrow d)_{PN} \ ; \\
      & \phantom{==}   (\emptyset \rightarrow a)_{PR} \ ; \ (\emptyset \rightarrow a)_{PK} \ ; \ (\emptyset \rightarrow a)_{RK} \ ; \ (\emptyset \rightarrow a)_{RPK} \ ; \  (acd \rightarrow abcd)_{P} \ ; \\
      & \phantom{==} (abd \rightarrow abcd)_{RPK} \ ; \ (abc \rightarrow abcd)_{RPK} \ ; \ (ad \rightarrow abcd)_{PNRKS} \ ; \\  
      & \phantom{==}   (ab \rightarrow abcd)_{RPKS} \ ; \ (abc \rightarrow abcd)_{RPKS} \ ; \ (abd \rightarrow abcd)_{PNRKS}\} 
  \end{align*}  
  We can also check that those relevant quasi-features with respect to $C$ are as follows:
  \begin{center}
 $ a{\times}P \ ; \ a{\times}N \ ; \ a{\times}R \ ; \ a{\times}K \ ; \ a{\times}S \ ; \  a{\times}PR \ ; \  a{\times}PK \ ; \  a{\times}RK \ ; \ b{\times}P \ ;\ b{\times}S \ ; \ c{\times}P $ \\ 
 $  c{\times}S \ ; \ c{\times}PS \ ; \ d{\times}P \ ; \ d{\times}N \ ; \ d{\times}R \ ; \ d{\times}K \ ; \ d{\times}S \ ; \  bc{\times}S \ ; \ a{\times} \emptyset \ ; \ d{\times} \emptyset \ ; \ ad{\times} \emptyset$ \ ; \\
 $acd{\times}P \ ; \ abd{\times}RPK \ ; \ abc{\times}RPK \ ; \ ab{\times}RPKS \ ; \ d{\times}PNK \ ; \ a{\times}PNRK \ ; \ d{\times}PNRK$ ; \\ $d{\times}PNRS \ ; \ d{\times}PNKS \ ; \ abc{\times}RPKS \ ; \ abcd{\times}P \ ; \ abcd{\times}PN \ ; \ abcd{\times}RPK.$
 \end{center}
   Therefore, 
   \begin{align*}
       \mathcal{B}_{BACI} &= \{(P \rightarrow KPR)_{a} \ ; \ (N \rightarrow KPNRS)_{a} \ ; \ (R \rightarrow KPR)_{a} \ ; \ (K \rightarrow KPR)_{a} \ ; \\
                          & \phantom{==} (S \rightarrow KRPS)_{a} \ ; \ (PR \rightarrow KPR)_{a} \ ; \ (PK \rightarrow KPR)_{a} \ ; \ (RK \rightarrow KPR)_{a} \ ; \\
                          & \phantom{==} (P \rightarrow KP)_{b} \ ; \ (S \rightarrow KPNRS)_{b} \ ; \ (P \rightarrow KPNRS)_{c} \ ; \ (S \rightarrow KPNRS)_{c} \ ; \\
                          & \phantom{==} (PS \rightarrow KPNRS)_{c} \ ; \ (P \rightarrow NP)_{d} \ ; \ (N \rightarrow NP)_{d} \ ; \ (R \rightarrow NPR)_{d} \ ; \\
                          & \phantom{==} (K \rightarrow KPNRS)_{d} \ ; \ (S \rightarrow NPS)_{d} \ ; \  (S \rightarrow KPNRS)_{bc} \ ; \ (\emptyset \rightarrow RPK)_{a} \ ; \\
                          & \phantom{==}  (\emptyset \rightarrow PN)_{d} \ ; \ (\emptyset \rightarrow P)_{ad} \ ; \ (P \rightarrow PNRKS)_{acd} \ ; \ (RPK \rightarrow PNRKS)_{abd} \ ; \\  
                          & \phantom{==} (RPK \rightarrow PNRKS)_{abc} \ ; \ (RPKS \rightarrow PNRKS)_{ab} \ ; \ (PNK \rightarrow PNRKS)_{d} \ ; \\ 
                          & \phantom{==} (PNRK \rightarrow PNRKS)_{a} \ ; \ (PNRK \rightarrow PNRKS)_{d} \ ; \ (PNRS \rightarrow PNRKS)_{d} \ ; \\
                          & \phantom{==} (PNKS \rightarrow PNRKS)_{d} \ ; \ (RPKS \rightarrow PNRKS)_{abc} \ ; \ (P \rightarrow PNRKS)_{abcd} \ ; \\
                          & \phantom{==}  (PN \rightarrow PNRKS)_{abcd} \ ; \ (RPK \rightarrow PNRKS)_{abcd}\}
   \end{align*}
   
\end{example}

   The absence of a relation between some attributes-conditions-objects in a context can be considered as information that can be described by implications. In what follows, we highlight a few remarkable facts about them. The third assertion describes a rule almost similar to \textbf{[Conditional decomposition]}.
\begin{proposition}\label{propo sur les implications a support nul}
   Let $K=(G,M,\mathcal{C},I)$ be our triadic context. For all $X,Y \subseteq M$ and $C \subseteq \mathcal{C}$, 
   \begin{enumerate}
       \item If $X^{(2,1,C)}=\emptyset$, then $(X \rightarrow M\backslash X)_C$ is valid in $\KK$.
       \item If $(Y\backslash X)^{(2,1,C)}=\emptyset$ and $(X \rightarrow Y)_C$ is valid in $\KK$, then $(Y \rightarrow X)_C$ is valid in $\KK$.
       \item If $(X \rightarrow Y)_C$ is valid in $\KK$ and $X^{(2,1,C_1)}=\emptyset$ with $C_1 \subseteq C$, then $(X \rightarrow Y)_{C_1}$ is valid in $\KK$.
   \end{enumerate}
\end{proposition}
\begin{proof}
    It is obvious that $(X \rightarrow M\backslash X)_C$ is valid in $\KK$ since $X^{(2,1,C)}=\emptyset$. As for $2$, if $(X \rightarrow Y)_C$ is valid in $\KK$, then $X^{(2,1,C)} \subseteq Y^{(2,1,C)} \subseteq (Y\backslash X)^{(2,1,C)} = \emptyset \subseteq X^{(2,1,C)}$, i.e. $Y^{(2,1,C)} \subseteq X^{(2,1,C)}$. So, $(Y \rightarrow X)_C$ is valid in $\KK$. Finally, the  third assertion follows from the first.
\end{proof}

 In the examples below, we deduce new implications from $\mathcal{B}_{BCAI}$. 
 \begin{multicols}{2}
 \textbf{Ex~1:}
 \begin{enumerate}
    \item $(ad \rightarrow M)_{\mathcal{C}}$ from $\mathcal{B}_{BCAI}$;
    \item $(ad \rightarrow bc)_{\mathcal{C}}$ from $1$ and \textbf{[Decomposition]};
    \item $(ad \rightarrow bc)_{S}$ from $1$ and Proposition~\ref{propo sur les implications a support nul}~3.
    \item $(bc \rightarrow ad)_{S}$ from $3$ and Proposition~\ref{propo sur les implications a support nul}~2.
\end{enumerate}
\textbf{Ex~2:}
 \begin{enumerate}
    \item $(abd \rightarrow M)_{RPK}$ and $(\emptyset \rightarrow a)_{RPK}$ from $\mathcal{B}_{BCAI}$;
    \item $(bd \rightarrow M)_{RPK}$ from $1$ and \textbf{[Pseudo-transitivity]}.
    \item $(bd \rightarrow M)_{K}$ from $3$ and Proposition~\ref{propo sur les implications a support nul}~3.
\end{enumerate}

\newcolumn
\textbf{Ex~3:}
\begin{enumerate}
    \item[a] $(c \rightarrow bd)_{N}, \ (d \rightarrow ab)_{R}$ from $\mathcal{B}_{BCAI}$;
    \item $(cd \rightarrow b)_{NR}$ from $1$ and \textbf{[Conditional composition]};
    \item $(bc \rightarrow ad)_{NR}$ from $2$ and Proposition~\ref{propo sur les implications a support nul}~1.;
    \item $(ad \rightarrow bc)_{S}$ from $3$ and Proposition~\ref{propo sur les implications a support nul}~2.;
    \item $(ad \rightarrow c)_{S}$ from $4$ and \textbf{[Decomposition]};
    \item $(c \rightarrow abcd)_{S}$ Proposition~\ref{propo sur les implications a support nul}~2.
\end{enumerate}

\end{multicols}

  With a complete set of implications $\Sigma$ of $\KK$, we are sure that all implications valid in $\KK$ can be derived from $\Sigma$. However, we cannot confirm whether there could exist some implications $\sigma_{i}$ in $\Sigma$ that could be derived from $\Sigma \setminus \{\sigma_{i}\}$. Thus, it is important to search for a minimum set of valid implications.

\begin{definition}
Let $\Sigma$ be a set of implications in $\KK$.
\begin{enumerate}
    \item $\Sigma$ is \textbf{non-redundant} if for all $\sigma$ in $\Sigma$, $\Sigma \setminus \{\sigma\} \nvdash \sigma$. 
    \item  A complete and non-redundant set of implications is called a \textbf{non-redundant base}~\cite{Belohlavek2008}.
\end{enumerate} 
\end{definition}

      To avoid redundancy, Sebastian Rudolph proved that an incrementor $P$ is a pseudo-intent if, for any incrementor $Q$, there exists a feature $R$ such that $Q \subseteq R \subseteq P$ \cite{Rudolph2007}.
  In triadic contexts, this can be adapted (with respect to simplification logic) as follows: for any relevant quasi-feature $A{\times}C$, 
  \begin{itemize}
      \item \textbf{[conditional composition]}: if there are some other quasi-features $A_i{\times}C_i$, strict subsets of $A \times C$, such that $\underset{i}{\cup} C_i = C$ and $A^{(1,2,C)(1,2,C)} = \underset{i}{\cap} A_i^{(1,2,C_i)(1,2,C_i)}$, then $\{ (A_i \rightarrow A_i^{(1,2,C_i)(1,2,C_i)} )_{_{C_i}} : \ i \in \{1, \dots, n\} \} \vdash (A \rightarrow A^{(1,2,C)(1,2,C)} )_{_C}.$
       \item \textbf{[Augmentation]}: if there is another quasi-feature $A_1{\times}C_1$, strict subsets of $A \times C$, such that $C_1 = C$ and $A^{(1,2,C)(1,2,C)} = A_1^{(2,1,C_1)(1,2,C_1)}$, then $(A_1 \rightarrow A_1^{(1,2,C_1)(1,2,C_1)} )_{_{C_1}} \vdash (A \rightarrow A^{(1,2,C)(1,2,C)} )_{_C}.$
       \item \textbf{[Transitivity]}: if there are some other quasi-features $A_i{\times}C_i$, such that $C_i = C, \ A_1=A, \ A_{n-1}^{(1,2,C_{n-1})(1,2,C_{n-1})}=A_n$ and $A^{(1,2,C)(1,2,C)} = A_n^{(1,2,C_n)(1,2,C_n)}$, then $\{ (A_i \rightarrow A_i^{(1,2,C_i)(1,2,C_i)} )_{_{C_i}} : \ i \in \{1, \dots, n\}\} \vdash (A \rightarrow A^{(1,2,C)(1,2,C)} )_{_C}.$
  \end{itemize} 
 
  To simplify the notation, we can write $\{A_i{\times}C_i:\ i \in \mathbb{N} \} \vdash A{\times}C$ instead of $$\{ (A_i \rightarrow A_i^{(1,2,C_i)(1,2,C_i)} )_{_{C_i}} : \ i \in \{1, \dots, n\}\} \vdash (A \rightarrow A^{(1,2,C)(1,2,C)} )_{_C}.$$
 
 \begin{definition}
  In a triadic context $\KK=(G, M, \mathcal{C}, I)$, the \textbf{minimal coverage of quasi-features} also called \textbf{pseudo-feature} with respect to $M$ noted $\mathcal{P}_2(\KK)$ (resp., with respect to $\mathcal{C}$ noted $\mathcal{P}_3(\KK)$), is the smallest set of all quasi-features of $\KK$ such that: for $i \in \{2,3\}$,     
  \begin{align*}
      & \star \ \mathcal{P}_i(\KK) \vdash A{\times}C, \ \text{for all quasi-feature} \ A{\times}C \\
      & \star \star \ \mathcal{P}_i(\KK)\backslash \{A{\times}C\} \nvdash A{\times}C
 \end{align*}
 \end{definition}\textbf{\\}
    In Example \ref{exple complete base}, we have 
    \begin{align*}
        & \emptyset {\times}P \vdash \{a{\times}P; \ b{\times}P; \ d{\times}P\}; \\
        & \emptyset {\times}N \vdash \{a{\times}N; \ b{\times}N\}; \\
        & \emptyset {\times}R \vdash \{b{\times}R; \ c{\times}R\}; \\
        & \emptyset {\times}K \vdash \{b{\times}K; \ d{\times}K\}; \\
        & \{\emptyset {\times}P; \ \emptyset {\times}N\} \vdash \emptyset {\times}PN; \\
        & \{\emptyset {\times}P; \ \emptyset {\times}R; \ \emptyset {\times}K\} \vdash \{ \emptyset {\times}PR; \ \emptyset {\times}PK; \ \emptyset {\times}KR; \ \emptyset {\times}RPK\}; \\
        & \{b{\times}S; \ c{\times}S\} \vdash  bc{\times}S; \\
        & \{c{\times}P; \ c{\times}S\} \vdash  c{\times}PS;  \\  
        & \{\emptyset {\times}P; \ c{\times}P\} \vdash acd{\times}P ; \\
        & \{ab{\times}RPKS; \ c{\times}S\} \vdash abc{\times}RPKS; \\
        &  ad{\times}PNRKS   \vdash  abd{\times}PNRKS.     
   \end{align*} Thus,

\begin{align*}
\mathcal{P}_2(\KK)=&\{\emptyset {\times}P \ ; \ \emptyset {\times}N \ ; \ \emptyset {\times}R \ ; \ \emptyset {\times}K \ ; \ c{\times}P \ ; \ d{\times}R \ ; \ c{\times}K \ ; \ c{\times}N \ ; \ b{\times}S \ ; \ c{\times}S \ ; \\
                   & abd{\times}RPK \ ; \ abc{\times}RPK \ ; \ ad{\times}PNRKS \ ; \ ab{\times}RPKS\}
   \end{align*}
Therefore, 
 \begin{align*}
     \mathcal{B}_{BCAI} & = \{(\emptyset \rightarrow ad)_{P} \ ; \ (\emptyset \rightarrow d)_{N} \ ; \ (\emptyset \rightarrow a)_{R} \ ; \ (\emptyset \rightarrow a)_{K} \ ; \ (c \rightarrow abcd)_{P} \ ; \ (d \rightarrow abd)_{R} \ ; \\ 
                       & \phantom{==} (c \rightarrow abc)_{K} \ ;  \ (c \rightarrow bcd)_{N} \ ;  \ (b \rightarrow abcd)_{S} \ ; \ (c \rightarrow abcd)_{S} \ ; \ (abd \rightarrow abcd)_{RPK} \ ; \\
                       & \phantom{==} (abc \rightarrow abcd)_{RPK} \ ; \ (ad \rightarrow abcd)_{PNRKS} \ ; \ (ab \rightarrow abcd)_{RPKS}\}
  \end{align*} 

Similarly, from the relevant quasi-features with respect to $C$ in Example \ref{exple complete base}, we have
    \begin{align*}
        & a{\times} \emptyset \vdash \{a{\times}P; \ a{\times}R; \ a{\times}K; \ a{\times}S; \  a{\times}PR; \  a{\times}PK; \  a{\times}RK \}; \\
        & d{\times} \emptyset \vdash \{d{\times}P; \ d{\times}N; \ d{\times}R; \ d{\times}S \}; \\
        & \{a{\times} \emptyset; \ d{\times} \emptyset\}  \vdash \{ad{\times} \emptyset \}; \\
        & \{c{\times}P; \ c{\times}S\} \vdash  c{\times}PS; \\
        & \{b{\times}S; \ c{\times}S\} \vdash  bc{\times}S;  \\
        & a{\times}N  \vdash a{\times}PNRK; \\
        &  d{\times}K \vdash \{ d{\times}PNK; \ d{\times}PNRK ; \ d{\times}PNKS\} \\
        & abcd{\times}P  \vdash  \{abcd{\times}PN; \ abcd{\times}RPK\}        \\
        &  abc{\times}RPK   \vdash  abc{\times}RPKS     \\
   \end{align*}

 Therefore,  
 \begin{align*}
  \mathcal{P}_3(\KK) &= \{ a{\times}N \ ; \ b{\times}P \ ;\ b{\times}S \ ; \ c{\times}P \ ; c{\times}S \ ; \ d{\times}K \ ; \ a{\times} \emptyset \ ; \ d{\times} \emptyset \ ; \ acd{\times}P \ ; \ abd{\times}PRK \ ; \\
                    & \phantom{==} abc{\times}PRK \ ; \ ab{\times}RPKS \ ; \ d{\times}PNRS \ ; \ abcd{\times}P \}.
 \end{align*}
 That is,
   \begin{align*}
       \mathcal{B}_{BACI} &= \{ (N \rightarrow KPNRS)_{a} \ ; \ (P \rightarrow KP)_{b} \ ; \ (S \rightarrow KPNRS)_{b} \ ; \  (P \rightarrow KPNRS)_{c} \ ; \\
                          & \phantom{==}   (S \rightarrow KPNRS)_{c} \ ; \ (K \rightarrow KPNRS)_{d} \ ; \ (\emptyset \rightarrow RPK)_{a} \ ; \ (\emptyset \rightarrow PN)_{d} \ ; \\ 
                          & \phantom{==}  (P \rightarrow PNRKS)_{acd} \ ; \ (RPK \rightarrow PNRKS)_{abd} \ ; \ (RPK \rightarrow PNRKS)_{abc} \ ; \\
                          & \phantom{==} (RPKS \rightarrow PNRKS)_{ab} \ ; \ (PNRS \rightarrow PNRKS)_{d} \ ; \ (P \rightarrow PNRKS)_{abcd}  \}       
   \end{align*} 

We then examine these bases of implications, paying particular attention to the minimal criterion.

\begin{definition}
    Let $\Sigma$ and $\Sigma_{1}$ be two sets of implications in $\KK$.
    \begin{enumerate}
        \item If $\Sigma \vdash \sigma$ for all $\sigma$ in $\Sigma_{1}$, then we can write $\Sigma \vdash \Sigma_{1}$; moreover, if $\Sigma_{1} \vdash \Sigma$ then the sets $\Sigma$ and $\Sigma_{1}$ are said to be \textbf{equivalent}.
        \item $\mathcal{B}$ is a base with a \textbf{minimum cardinality} when $\mathcal{B}$ is a base and for any complete set of implications $\mathcal{B}_{1}$, it holds $|\mathcal{B}|\leq|\mathcal{B}_{1}|$. That is, $\mathcal{B}$ has as few implications as any equivalent set of implications. Furthermore, $\mathcal{B}$ is a \textbf{minimal base} if $\mathcal{B}$ is a base satisfying $$\forall \sigma \in \mathcal{B}, \quad \mathcal{B} \setminus \{\sigma\} \nvdash \mathcal{B}$$
   \end{enumerate}
\end{definition}


\begin{lemma}\label{lemme minimal base}\cite{RomualdBlaiseLeonardEtienne2025}
 If $\Sigma$ is a complete set of BCAI (respectively BACI) of $\KK$, then for each relevant quasi-feature $X{\times}Y \in \mathcal{P}_{2}(\KK)$ (respectively, $X{\times}Y \in \mathcal{P}_{3}(\KK)$),
  $\Sigma$ contains an implication $\sigma=(A_{1} \rightarrow A_{2})_{C}$ such that $A_{1}^{(1,2,C)(1,2,C)} = X^{(1,2,Y)(1,2,Y)}$ (respectively, an implication $\sigma=(C_{1} \rightarrow C_{2})_{A}$ such that $C_{1}^{(1,3,A)(1,3,A)} = Y^{(1,3,X)(1,3,X)}$).
\end{lemma}
\begin{proof}

Without loss of generality, assume that $X{\times}Y$ is a relevant quasi-feature with respect to $M$ and $\Sigma$ a complete set of BCAI. We have $\Sigma \vdash (X \rightarrow X^{(1,2,Y)(1,2,Y)})_{_Y}$, that is,  $X^{(1,2,Y)(1,2,Y)}\subseteq (X)_{\Sigma,Y}$. As $X^{(1,2,Y)(1,2,Y)}$ is an intent, $X^{(1,2,Y)(1,2,Y)}=(X)_{\Sigma,Y}$. Finally, $\Sigma$ is complete means that it contains at least one implication of the form $(X_1\rightarrow W)_{_Y}$ with $X_1 \subseteq X$ and $W \subseteq (X)_{\Sigma,Y}$. 
\end{proof}
\begin{theorem}\label{theo minimum base of BCAI}\cite{RomualdBlaiseLeonardEtienne2025}
In a triadic context $\KK$, the sets $\mathcal{B}_{BCAI}$ and $\mathcal{B}_{BACI}$ are minimal bases of implications.
\begin{eqnarray*}
 \mathcal{B}_{BCAI} = \{ (A \rightarrow A^{(1,2,C)(1,2,C)} )_{_C} : A{\times}C \in \mathcal{P}_{2}(\KK) \}  \\
 \mathcal{B}_{BACI} = \{ (C \rightarrow C^{(1,3,A)(1,3,A)} )_{_A} : A{\times}C \in \mathcal{P}_{3}(\KK) \}  
\end{eqnarray*}
\end{theorem}

\begin{proof}
According to Lemma~\ref{lemme valid implication}, all implications in $\mathcal{B}_{BCAI}$ are valid in the context $\KK$. Therefore, Lemma~\ref{lemme complete set of implications} reassures us that $\mathcal{B}_{BCAI}$ is complete. The non-redundancy comes from the definition of $\mathcal{P}_{2}(\KK)$. The minimality of $\mathcal{B}_{BCAI}$ follows from the definition of $\mathcal{P}_{2}(\KK)$ and Lemma~\ref{lemme minimal base}. Thus, $\mathcal{B}_{BCAI}$ is a minimal base of implications.

The proof is similar for BACI.


\end{proof}

In the following paragraph, we want to construct a base for CAI and ACI. To do this, we begin by recalling the interplay between an implication of the form $A_1 \overset{C}{\rightarrow} A_2$ and that of the form $(A_1 \rightarrow A_2)_{C}$ as described below: 
\begin{equation}\label{equivalence CAI et BCAI}
    A_1 \overset{C}{\rightarrow} A_2 \quad \text{if and only if} \quad (A_1 \rightarrow A_2)_{\{c\}}, \quad \forall c \in C 
\end{equation}
This relationship is called (unary) \textbf{conditional decomposition} (see \cite{RodriguezEncisoRokiaMora2017,RodriguezCorderoRokiaMora2016} for more details). We can see that an implication of the form $(A_1 \rightarrow A_2)_{\{c\}}$ can be deduced from a relevant quasi-feature of the form $A{\times}\{c\}$ call a \textbf{relevant unit quasi-feature of $\KK$ with respect to $M$}. In what follows, we will note $\mathbb{UP}_2(\mathbb{K})$ (respectively, $\mathbb{UP}_3(\mathbb{K})$) the set of relevant unit quasi-features of $\KK$ with respect to $M$ (respectively, $\mathcal{C}$). 
\begin{remark}\label{base of CAI}
     In \cite{GanterObiedkov2004} P. 187, the author confirms that the definition \textbf{"$A \overset{C}{\rightarrow} B$ is valid in a context $\KK$ if and only if $(A \rightarrow B)_c$, for all $c\in C$ are valid in the same context"} is interrelated with the definition \textbf{"$A \overset{C}{\rightarrow}B$ is valid in a context $\KK$ if and only if $(A \rightarrow B)_X$, for all $X\subseteq C$ are valid in the same context"}. Furthermore, any base $\mathcal{B}= \{A \overset{C}{\rightarrow} B \ / \ A, B \subseteq M;\ C \subseteq \mathcal{C}\}$ of CAI corresponds to the set $T =\{(A \rightarrow B)_{c}, \ c \in C \ / \ A \overset{C}{\rightarrow} B \in \mathcal{B}\}$ which is not necessarily complete and can be generated by $\mathcal{B}_{BCAI}$. As conditional decomposition is not possible for BCAI, $T$ will only be deduced from $\mathbb{UP}_2(\mathbb{K})$. Therefore, the set $\mathbb{UP}_2(\mathbb{K})$ alone is sufficient to generate a CAI base. The following theorem states this fact.
\end{remark}

\begin{theorem}\label{theorem base of CAI} 
The following sets of implications are bases in $\KK$.
\begin{eqnarray*}
 \mathcal{B}_{CAI} = \{A \overset{c}{\rightarrow} A^{(1,2,c)(1,2,c)}: \ A{\times}c \in \mathbb{UP}_{2}(\KK) \}  \\
 \mathcal{B}_{ACI} = \{C \overset{a}{\rightarrow} C^{(1,3,a)(1,3,a)}: \ C{\times}a \in \mathbb{UP}_{3}(\KK) \}  
\end{eqnarray*}
\end{theorem}
\begin{proof}
 The validity and relevance of each implication follow from Lemma~\ref{lemme valid implication} and the fact that these unit quasi-features are relevant. The completeness of the constructed sets is a corollary of Remark~\ref{base of CAI}.
\end{proof}
\begin{remark}
The bases $\mathcal{B}_{CAI}$ and $\mathcal{B}_{ACI}$ can be made minimal using the simplification logic. We present an illustration of this statement for CAI in the following example.   
\end{remark}
\begin{example}\label{exple base}
 From Example~\ref{exple complete base},
 \begin{align*}
   \mathbb{UP}_2(\KK)= & \{a{\times}P \ ; \ a{\times}N \ ; \ b{\times}P \ ;\ b{\times}N \ ; \  b{\times}R \ ; \  b{\times}K \ ; \ b{\times}S \ ; \ c{\times}P \ ;\ c{\times}N \ ; \  c{\times}R \ ; \\
                    & \phantom{==} c{\times}K \ ; \ c{\times}S \ ; \ d{\times}P \ ; \ d{\times}R \ ; \ d{\times}K \ ; \ bc{\times}S \ ; \  \emptyset {\times} P \ ; \  \emptyset {\times} N \ ; \\
                    & \phantom{==} \emptyset {\times} R \ ; \  \emptyset {\times} K \}
 \end{align*}
Therefore, we can deduce that:
  \begin{align*}
     \mathcal{B}_{CAI} & = \{a \overset{P}{\rightarrow} ad \ ; \ a \overset{N}{\rightarrow} ad \ ;  \ b \overset{P}{\rightarrow} abd \ ; \ b \overset{N}{\rightarrow} bd \ ;  \ b \overset{R}{\rightarrow} ab \ ; \ b \overset{K}{\rightarrow} ab ; \  b \overset{S}{\rightarrow} abcd \ ; \\ 
      & \phantom{==}    c \overset{P}{\rightarrow} abcd \ ; \ c \overset{N}{\rightarrow} bcd \ ;  \ c \overset{R}{\rightarrow} ac \ ; \ c \overset{K}{\rightarrow} abc \ ; \ c \overset{S}{\rightarrow} abcd \ ; \ d \overset{P}{\rightarrow} ad \ ; \ d \overset{R}{\rightarrow} abd \ ; \\
      & \phantom{==}   d \overset{K}{\rightarrow} ad \ ; \ bc \overset{S}{\rightarrow} abcd \ ; \ \emptyset \overset{P}{\rightarrow} ad \ ; \ \emptyset \overset{N}{\rightarrow} d \ ; \ \emptyset \overset{R}{\rightarrow} a \ ; \ \emptyset \overset{K}{\rightarrow} a \} 
  \end{align*} 
  Since we have $\{\emptyset \overset{P}{\rightarrow} ad \ ; \ \emptyset \overset{N}{\rightarrow} d \ ; \ \emptyset \overset{R}{\rightarrow} a \ ; \ \emptyset \overset{K}{\rightarrow} a\} \vdash\{a \overset{P}{\rightarrow} ad \ ; \ a \overset{N}{\rightarrow} ad \ ;  \ b \overset{P}{\rightarrow} abd \ ; \ b \overset{N}{\rightarrow} bd \ ;  \ b \overset{R}{\rightarrow} ab \ ; \ b \overset{K}{\rightarrow} ab ; \ c \overset{R}{\rightarrow} ac ; \ d \overset{P}{\rightarrow} ad \ ; \ d \overset{K}{\rightarrow} ad \}$, the set $\mathcal{B}_{CAI}$ can be reduced to form
  \begin{align*}
     \mathcal{B}_{CAI} & = \{b \overset{S}{\rightarrow} abcd \ ; \ c \overset{P}{\rightarrow} abcd \ ; \ c \overset{N}{\rightarrow} bcd \ ; \ c \overset{K}{\rightarrow} abc \ ; \ c \overset{S}{\rightarrow} abcd \ ; \ d \overset{R}{\rightarrow} abd \ ; \\
      & \phantom{==}    bc \overset{S}{\rightarrow} abcd \ ; \ \emptyset \overset{P}{\rightarrow} ad \ ; \ \emptyset \overset{N}{\rightarrow} d \ ; \ \emptyset \overset{R}{\rightarrow} a \ ; \ \emptyset \overset{K}{\rightarrow} a \} 
  \end{align*}
 Similarly, $\{b \overset{S}{\rightarrow} abcd \ ; \ c \overset{S}{\rightarrow} abcd\} \vdash bc \overset{S}{\rightarrow} abcd$. Deleting $bc \overset{S}{\rightarrow} abcd$ in $\mathcal{B}_{CAI}$ gives rise to the minimal base of CAI
  \begin{align*}
     \mathcal{B}_{CAI} & = \{b \overset{S}{\rightarrow} abcd \ ; \ c \overset{P}{\rightarrow} abcd \ ; \ c \overset{N}{\rightarrow} bcd \ ; \ c \overset{K}{\rightarrow} abc \ ; \ c \overset{S}{\rightarrow} abcd \ ; \ d \overset{R}{\rightarrow} abd \ ; \\
      & \phantom{==}  \emptyset \overset{P}{\rightarrow} ad \ ; \ \emptyset \overset{N}{\rightarrow} d \ ; \ \emptyset \overset{R}{\rightarrow} a \ ; \ \emptyset \overset{K}{\rightarrow} a \} 
  \end{align*}

\begin{remark}
    We can verify that this minimality follows from the fact that the set $\mathcal{B}_{CAI}$ can be obtained directly from the minimal coverage of unit quasi-features i.e., pseudo-features. 
\end{remark}
  
  
  We can also check that those relevant unit quasi-features with respect to $C$ are as follows:
   \begin{align*}
  \mathbb{UP}_3(\KK)= & \{a{\times}P \ ; \ a{\times}N \ ; \ a{\times}R \ ; \ a{\times}K \ ; \ a{\times}S \ ; \  a{\times}PR \ ; \  a{\times}PK \ ; \  a{\times}RK \ ; \ b{\times}P \ ; \\
                      & \phantom{==} b{\times}S \ ; \ c{\times}P \ ; \ c{\times}S \ ; \ c{\times}PS \ ; \ d{\times}P \ ; \ d{\times}N \ ; \ d{\times}R \ ; \ d{\times}K \ ; \ d{\times}S \ ; \ a{\times} \emptyset \ ; \\
                      & \phantom{==} d{\times} \emptyset \ ; \ d{\times}PNK \ ;  \ a{\times}PNRK \ ; \ d{\times}PNRK \ ; \ d{\times}PNRS \ ; \ d{\times}PNKS.\}
 \end{align*}
   Therefore, 
   \begin{align*}
       \mathcal{B}_{ACI} &= \{P \overset{a}{\rightarrow} KPR \ ; \ N \overset{a}{\rightarrow} KPNRS \ ; \ R \overset{a}{\rightarrow} KPR \ ; \ K \overset{a}{\rightarrow} KPR \ ; \\
                          & \phantom{==} S \overset{a}{\rightarrow} KRPS \ ; \ PR \overset{a}{\rightarrow} KPR \ ; \ PK \overset{a}{\rightarrow} KPR \ ; \ RK \overset{a}{\rightarrow} KPR \ ; \\
                          & \phantom{==} P \overset{b}{\rightarrow} KP \ ; \ S \overset{b}{\rightarrow} KPNRS \ ; \ P \overset{c}{\rightarrow} KPNRS \ ; \ S \overset{c}{\rightarrow} KPNRS \ ; \\
                          & \phantom{==} PS \overset{c}{\rightarrow} KPNRS \ ; \ P \overset{d}{\rightarrow} NP \ ; \ N \overset{d}{\rightarrow} NP \ ; \ R \overset{d}{\rightarrow} NPR \ ; \\
                          & \phantom{==} K \overset{d}{\rightarrow} KPNRS \ ; \ S \overset{d}{\rightarrow} NPS \ ; \ \emptyset \overset{a}{\rightarrow} RPK \ ; \ \emptyset \overset{d}{\rightarrow} PN \ ; \\
                          & \phantom{==} PNK \overset{d}{\rightarrow} PNRKS \ ; \ PNRK \overset{a}{\rightarrow} PNRKS \ ; \ PNRK \overset{d}{\rightarrow} PNRKS \ ; \\
                          & \phantom{==}  PNRS \overset{d}{\rightarrow} PNRKS \ ; \ PNKS \overset{d}{\rightarrow} PNRKS\}
   \end{align*}
 
   Considering the implication $P \overset{abc}{\rightarrow} K$ valid in the context of Fig.~\ref{tab 1} (right) (see Example~\ref{example 8}), we want to illustrate the proof of the assertion $\mathcal{B}_{ACI} \vdash P \overset{abc}{\rightarrow} K$ with the following derivation sequences. 
   \begin{enumerate}
       \item $P \overset{a}{\rightarrow} KPR$, belongs to $\mathcal{B}_{ACI}$;
       \item $P \overset{b}{\rightarrow} KP$, belongs to $\mathcal{B}_{ACI}$;
       \item $P \overset{c}{\rightarrow} KPNRS$, belongs to $\mathcal{B}_{ACI}$;
       \item $P \overset{abc}{\rightarrow} KP$, from 1,2 3 and \textbf{[Conditional composition]};
       \item $P \overset{abc}{\rightarrow} K$, from \textbf{[Decomposition]}.
   \end{enumerate}
 \end{example} 

 We have evaluated our bases according to the number of implications they contain. We can also evaluate them by the number of elements (different or not) present in each implication. This number is the size of the base. For example, the size of $\{(a \rightarrow ad)_{P}\}$ is 4 and the set $\{(a \rightarrow ad)_{P} \ ; \ (\emptyset \rightarrow d)_{N} \ ; \ (\emptyset \rightarrow a)_{R}\}$ has size 8.
\begin{definition}\cite{RomualdBlaiseLeonardEtienne2025}
 A set $\mathcal{B}$ of implications of a context $\KK$, is an \textbf{optimal base} when $\mathcal{B}$ is a base and for any equivalent set of implications $\mathcal{B}_1$, we have $\|\mathcal{B}\| \leq \|\mathcal{B}_1\|$, where $ \|\mathcal{B}\| := \underset{(A \rightarrow B)_C \in \mathcal{B}}\sum (|A|+|B|+|C|)$.
\end{definition}

   From \cite{RomualdBlaiseLeonardEtienne2025}, we know that the optimal base of BCAI in our running context is
\begin{align*}
     \mathcal{B}^{op}_{BCAI} & = \{(\emptyset \rightarrow ad)_{P} \ ; \ (\emptyset \rightarrow d)_{N} \ ; \ (\emptyset \rightarrow a)_{R} \ ; \ (\emptyset \rightarrow a)_{K} \ ; \ (c \rightarrow b)_{P} \ ; \ (d \rightarrow b)_{R} \ ; \\ 
                       & \phantom{==} (c \rightarrow ab)_{K} \ ;  \ (c \rightarrow b)_{N} \ ;  \ (b \rightarrow acd)_{S} \ ; \ (c \rightarrow abd)_{S} \} 
  \end{align*} 
 
We now want to determine the optimal base for CAI and ACI, respectively.

 In what follows, we will note $\mathcal{UP}_2(\mathbb{K})$ (respectively, $\mathcal{UP}_3(\mathbb{K})$) the unit pseudo-features with respect to $M$ (respectively, $\mathcal{P}_3(\mathbb{K})$ the unit pseudo-features with respect to $\mathcal{C}$).
 
We have the following result.

\begin{theorem}\label{theorem optimal base of CAI}
With respect to the logic of simplification of implications, the following bases are minimal and can be optimal in $\KK$.
\begin{eqnarray*}
 \mathcal{B}_{CAI}^{op} = \{ A \overset{c}{\rightarrow} A^{(1,2,c)(1,2,c)}\backslash A : A{\times}c \in \mathcal{UP}_{2}(\KK) \}  \\
 \mathcal{B}_{ACI}^{op} = \{ C \overset{a}{\rightarrow} C^{(1,3,a)(1,3,a)}\backslash C : a{\times}C \in \mathcal{UP}_{3}(\KK)\}  
\end{eqnarray*}
\end{theorem}
\begin{proof}
From Theorem~\ref{theorem base of CAI}, $\mathcal{B}_{CAI}^{op}$ is a base. Furthermore, $\mathcal{B}_{CAI}^{op}$ is built from unit pseudo-features $\mathcal{UP}_2(\mathbb{K})$ with respect to $M$; so, $\mathcal{B}_{CAI}^{op}$ is minimal. Moreover, the simplification logic and the reduction due to the implications described in Proposition~\ref{propo 100percent-support implication and optimality} guarantee optimality.


    The proof is similar for $\mathcal{B}_{ACI}^{op}$.
\end{proof}

 In the following examples, we compute the optimal bases for CAI and ACI in our running context. Achieving a $70\%$ reduction rate, we moved from $\mathcal{B}_{CAI}$ and $\mathcal{B}_{ACI}$ base to their respective optimal bases ($\mathcal{B}_{CAI}^{op}$ and $\mathcal{B}_{ACI}^{op}$).
\begin{example}\label{exple optimalbase CAI}
  In this example, we will compute the set $\mathcal{B}_{CAI}^{op}$ of the context in Fig.~\ref{tab 1} (left). We already know that 
 \begin{align*}
\mathcal{P}_2(\KK)=&\{\emptyset {\times}P \ ; \ \emptyset {\times}N \ ; \ \emptyset {\times}R \ ; \ \emptyset {\times}K \ ; \ c{\times}P \ ; \ d{\times}R \ ; \ c{\times}K \ ; \ c{\times}N \ ; \ b{\times}S \ ; \ c{\times}S \ ; \\
                   & abd{\times}RPK \ ; \ abc{\times}RPK \ ; \ ad{\times}PNRKS \ ; \ ab{\times}RPKS\}
   \end{align*}
   So, 
  \begin{align*}
\mathcal{UP}_2(\KK)=&\{\emptyset {\times}P \ ; \ \emptyset {\times}N \ ; \ \emptyset {\times}R \ ; \ \emptyset {\times}K \ ; \ c{\times}P \ ; \ d{\times}R \ ; \ c{\times}K \ ; \ c{\times}N \ ; \ b{\times}S \ ; \ c{\times}S \}.
   \end{align*}
  The following base is built from $\mathcal{UP}_2(\KK)$:
    $\{\emptyset \overset{P}{\rightarrow} ad \ ; \ \emptyset \overset{N}{\rightarrow} d \ ; \ \emptyset \overset{R}{\rightarrow} a \ ; \ \emptyset \overset{K}{\rightarrow} a \ ; \ c \overset{P}{\rightarrow} abd \ ; \ d \overset{R}{\rightarrow} ab \ ; \ c \overset{K}{\rightarrow} ab \ ;  \ c \overset{N}{\rightarrow} bd \ ;  \ b \overset{S}{\rightarrow} acd \ ; \  c \overset{S}{\rightarrow} abd \}$
  Since $\{\emptyset \overset{P}{\rightarrow} ad \ ; \ c \overset{P}{\rightarrow} b\} \vdash c \overset{P}{\rightarrow} abd$ from \textbf{[Accumulation]}, $c \overset{P}{\rightarrow} abd$ will undergo the \textbf{[Decomposition]} rule to form $c \overset{P}{\rightarrow} b$. We can also apply this simplification on $d \overset{R}{\rightarrow} ab \ ; \ c \overset{K}{\rightarrow} ab \ ;  \ c \overset{N}{\rightarrow} bd$ to obtain $d \overset{R}{\rightarrow} b \ ; \ c \overset{K}{\rightarrow} b \ ;  \ c \overset{N}{\rightarrow} b$, respectively 
  . The resulting base is 
   \begin{align*}
      \{\emptyset \overset{P}{\rightarrow} ad  ; \ \emptyset \overset{N}{\rightarrow} d  ; \ \emptyset \overset{R}{\rightarrow} a  ; \ \emptyset \overset{K}{\rightarrow} a  ; \ c \overset{P}{\rightarrow} b  ; \ d \overset{R}{\rightarrow} b  ; \ c \overset{K}{\rightarrow} b  ;  \ c \overset{N}{\rightarrow} b  ;  \ b \overset{S}{\rightarrow} acd  ; \  c \overset{S}{\rightarrow} abd\} 
  \end{align*}
 From \textbf{[Conditional composition]}, we have 
  \begin{align*}
      \{\emptyset \overset{P}{\rightarrow} ad \ ; \ \emptyset \overset{N}{\rightarrow} d \ ; \ \emptyset \overset{KR}{\rightarrow} a \ ; \ c \overset{KPN}{\longrightarrow} b \ ; \ d \overset{R}{\rightarrow} b \ ; \ b \overset{S}{\rightarrow} acd \ ; \  c \overset{S}{\rightarrow} abd\} 
  \end{align*}
 We then apply \textbf{[Accumulation]} to observe that $\{b \overset{S}{\rightarrow} c \ ; \  c \overset{S}{\rightarrow} abd \} \vdash b \overset{S}{\rightarrow} acd$, i.e. $b \overset{S}{\rightarrow} acd$ will undergo \textbf{[Decomposition]} to form $b \overset{S}{\rightarrow} c $. Finally, the optimal base with respect to $M$ is 
  \begin{align*}
     \mathcal{B}_{CAI}^{op} & = \{\emptyset \overset{P}{\rightarrow} ad \ ; \ \emptyset \overset{N}{\rightarrow} d \ ; \ \emptyset \overset{KR}{\rightarrow} a \ ; \  c \overset{KPN}{\longrightarrow} b \ ; \ d \overset{R}{\rightarrow} b \ ; \ b \overset{S}{\rightarrow} c \ ; \  c \overset{S}{\rightarrow} abd\}.
  \end{align*}
  We can verify that $|\mathcal{B}_{CAI}^{op}|=7<|\mathcal{B}_{CAI}|=20$ and $\|\mathcal{B}_{CAI}^{op}\|=24<\|\mathcal{B}_{CAI}\|=86$. Hence a reduction rate of $65\%$ and $72\%$ respectively for cardinality and the size.
 \end{example} 
 
\begin{example}
Now, we compute the set $\mathcal{B}_{ACI}^{op}$ of the right context in Fig.~\ref{tab 1}. We have seen that:
 \begin{align*}
  \mathcal{P}_3(\KK) &= \{ a{\times}N \ ; \ b{\times}P \ ;\ b{\times}S \ ; \ c{\times}P \ ; c{\times}S \ ; \ d{\times}K \ ; \ a{\times} \emptyset \ ; \ d{\times} \emptyset \ ; \ acd{\times}P \ ; \ abd{\times}PRK \ ; \\
                    & \phantom{==} abc{\times}PRK \ ; \ ab{\times}RPKS \ ; \ d{\times}PNRS \ ; \ abcd{\times}P \}.
 \end{align*} So,
\begin{align*}
  \mathcal{UP}_3(\KK)=& \{ a{\times}N \ ; \ b{\times}P \ ;\ b{\times}S \ ; \ c{\times}P \ ; c{\times}S \ ; \ d{\times}K \ ; \ a{\times} \emptyset \ ; \ d{\times} \emptyset \ ; \  d{\times}PNRS\}.
 \end{align*}
 The base resulting from it is:
   \begin{align*}
       & \{ N \overset{a}{\rightarrow} KPRS \ ; \ P \overset{b}{\rightarrow} K \ ; \ S \overset{b}{\rightarrow} KPNR \ ; \ P \overset{c}{\rightarrow} KNRS \ ; \ S \overset{c}{\rightarrow} KPNR \ ; \\
       & \phantom{==}  K \overset{d}{\rightarrow} PNRS \ ; \ \emptyset \overset{a}{\rightarrow} RPK \ ; \ \emptyset \overset{d}{\rightarrow} PN \ ; \ PNRS \overset{d}{\rightarrow} K\}
   \end{align*} 
   Since $\{\emptyset \overset{a}{\rightarrow} RPK  \ ; \ N \overset{a}{\rightarrow} S\} \vdash N \overset{a}{\rightarrow} KRPS$ and $\{\emptyset \overset{d}{\rightarrow} PN  \ ; \ K \overset{d}{\rightarrow} SR\} \vdash K \overset{d}{\rightarrow} PNRS$ hold from \textbf{[Accumulation]} and that $\{\emptyset \overset{d}{\rightarrow} PN  \ ; \ RS \overset{d}{\rightarrow} K\} \vdash PNRS \overset{d}{\rightarrow} K$ holds from \textbf{[Simplification]} (see \cite{RodriguezEncisoRokiaMora2017, RodriguezCorderoRokiaMora2016} for more details about the simplification rule), $N \overset{a}{\rightarrow} KPRS$, $K \overset{d}{\rightarrow} PNRS$ will undergo the \textbf{[Decomposition]} rule to form $N \overset{a}{\rightarrow} S$ and $K \overset{d}{\rightarrow} SR$, respectively, while  $PNRS \overset{d}{\rightarrow} K$ will be replace by $RS \overset{d}{\rightarrow} K$.  The resulting base is 
    \begin{align*}
       & \{N \overset{a}{\rightarrow} S \ ; \ P \overset{b}{\rightarrow} K \ ; \ S \overset{b}{\rightarrow} KPNR \ ; \ P \overset{c}{\rightarrow} KNRS \ ; \ S \overset{c}{\rightarrow} KPNR \ ; \\
       & \phantom{==}  K \overset{d}{\rightarrow} SR \ ; \ \emptyset \overset{a}{\rightarrow} RPK \ ; \ \emptyset \overset{d}{\rightarrow} PN \ ; \ RS \overset{d}{\rightarrow} K\}
   \end{align*} 
    Finally, we apply \textbf{[Conditional composition]}, to have the following optimal base of $\KK$ with respect to $\mathcal{C}$. 
   \begin{align*}
       \mathcal{B}_{ACI}^{op} = & \{N \overset{a}{\rightarrow} S \ ; \ P \overset{b}{\rightarrow} K \ ; \ S \overset{bc}{\rightarrow} KPNR \ ; \ P \overset{c}{\rightarrow} KNRS \ ; \\
                                & \phantom{==} K \overset{d}{\rightarrow} SR \ ; \ \emptyset \overset{a}{\rightarrow} RPK \ ; \ \emptyset \overset{d}{\rightarrow} PN \ ; \ RS \overset{d}{\rightarrow} K\}
   \end{align*} 
  With this optimal base, we can noticed that: $|\mathcal{B}_{ACI}^{op}|=8<|\mathcal{B}_{ACI}|=25$ and $\|\mathcal{B}_{ACI}^{op}\|=34<\|\mathcal{B}_{ACI}\|=160$. Hence a reduction rate of approximately $68\%$ and $78\%$ respectively for cardinality and the size.
 \end{example} 

 \begin{remark}
  An element is \textbf{extraneous} in an implication $\sigma$ of a base $\mathcal{B}$, if deleting this element in $\sigma$ produces a base $\mathcal{B}_1$ equivalent to $\mathcal{B}$. If all extraneous elements are deleted in the premise, $\sigma$ will be called \textbf{left-reduced} or \textbf{full}; if there are deleted in the conclusion,  $\sigma$ will be called \textbf{right-reduced}. In case the two reductions hold and the conclusion is not empty, $\sigma$ will be called \textbf{reduced}. A base of full/right-reduced/reduced implications is called full/right-reduced/reduced respectively. A base is call \textbf{canonical} if it is a base containing full implications with a singleton as conclusion (see \url{https://web.cecs.pdx.edu/~maier/TheoryBook/TRD.html} for more details). 

     Clearly, the optimal base $\mathcal{B}_{CAI}^{op}$ is reduced and minimal, but it is not canonical since the conclusions are not always singletons. However, we can generate the canonical base from the optimal base.
 \end{remark}

\section{Algorithm description}\label{algorithm description}

  In the following, we will outline an algorithm for calculating the sets $\mathcal{UP}_{2}(\KK)$ and $\mathcal{UP}_{3}(\KK)$. Since they are unit pseudo-feature and features of the augmented context,  we can restrict the search to $2^{M}{\times}\mathcal{C}$ (respectively, $M{\times}2^{\mathcal{C}}$). The features of the current context $\KK$, i.e. $\mathfrak{F}(\KK)$, are not taken into account, the study will finally be done on $\mathcal{N} = (2^{M}{\times}\mathcal{C})- \mathfrak{F}(\KK)$ (respectively, $\mathcal{N} = (M{\times}2^{\mathcal{C}})- \mathfrak{F}(\KK)$).
  
    The following algorithm runs through all products $ Z \in \mathcal{N}$ in search of all relevant quasi-features with respect to $M$. Note that the chosen ones will constitute the set $\mathbb{UP}_{2}(\KK)$. It takes as input the context $\KK$, the set $\mathcal{N}$ and all features of $\KK$, i.e. $\mathfrak{F}(\KK)$ and returns $\mathbb{UP}_{2}(\KK)$ the set of all relevant quasi-features with respect to $M$.
 \begin{description}
   \item[\textbf{Function} \ $pseudofeat_{2}(\KK, \mathcal{N}, \mathfrak{F}(\KK))$] :  
   \item[\textbf{Input}] The context $\KK$, $\mathcal{N}$ and $\mathfrak{F}(\KK)$. \\
    
     \begin{tabular}{l}
      $\mathbb{UP}_{2}(\KK) = [] $  \quad "" create a table to store all pseudo features of $\KK$ "" \\
      i=0   \qquad  "" initializing i to 0 ""   \\
          \textbf{For all} $ Z \in \mathcal{N}$ \\
          \qquad \begin{tabular}{l}
              \textbf{If} $ feat_{2}(Z,\KK, \mathfrak{F}(\KK))$ is true      \\
                \quad \begin{tabular}{l}
                    Add Z in the case number i of the table $\mathbb{P}_{2}(\KK)$    \\
                     $ i \leftarrow i+1 $  \\
                        \end{tabular}
                 \end{tabular}
         \end{tabular}
       \item[\textbf{Return}] $\mathbb{UP}_{2}(\KK)$
 \end{description}
  
  The function  \textcolor{blue}{$ feat_{2}(Z,\KK, \mathfrak{F}(\KK))$} verifies if the set $ Z=A{\times}c \in \mathcal{N}$ is a relevant quasi-feature of $\KK$ with respect to $M$. It takes as input the context $\KK$, the set $Z=A{\times}c$, and all features of $\KK$, i.e. $\mathfrak{F}(\KK)$ and returns "true" if Z is a relevant quasi-feature of $\KK$ with respect to $M$ and "false" if not.
 \begin{description}
    \item[\textbf{Function}]   $ feat_{2}(Z,\KK, \mathfrak{F}(\KK))$ :  \\
    \qquad   \begin{tabular}{l}
    
        Construct $ \mathfrak{F}(\KK[Z])$  \\
        $ \mathcal{D}(\KK) \leftarrow \mathfrak{F}(\KK[Z]) \setminus \mathfrak{F}(\KK)$  \\
          \textbf{If} $ \mathcal{D}(\KK) \neq \{ Z \}$, \\            
          \quad  \begin{tabular}{l}
               \textbf{return} "false" and terminate.                              \quad  \quad ""if true, $Z$ is already a quasi-feature"" \\
               \textbf{Else, if} $(A)^{(1,2,c)(2,1,c)} \backslash A$ is empty      \quad   \quad ""verifying if $Z$ is relevant"" \\         
                    \qquad  \begin{tabular}{l}
                       \textbf{return} "false" and terminate.   \\
                     \begin{tabular}{l}
                        \textbf{Else, Return} "true" and terminate.
                       \end{tabular}
                     \end{tabular}
               \end{tabular}
            \end{tabular}
 \end{description}
  The following algorithm constructs pseudo-features.

  \begin{description}
  \item[\textbf{Function} \ $mincover_{2}(\KK, X)$] :  
  \item[\textbf{Input}] The set $X \leftarrow pseudofeat_{2}(\KK, \mathcal{N}, \mathfrak{F}(\KK))$ and the context $\KK$.  \\
  
     \begin{tabular}{l}
      $X_1, store \leftarrow \emptyset$   \\
      \textbf{For all} $A{\times}C \in X$ \\
             \qquad \begin{tabular}{l}
              \textbf{If} $|A| = 0$      \\
                   \quad $store \leftarrow A{\times}C$, \qquad \quad ""$store$ has all quasi-features with one empty\\
                 \begin{tabular}{l}
              \textbf{If} $|A| > 1$                       \qquad \qquad \qquad  component""  \\
                   \quad $X_1 \leftarrow A{\times}C$, \\
                 \end{tabular}
                 \end{tabular}
         \end{tabular}\textbf{\\}
         $ X \leftarrow X \backslash store$, \\
         \textbf{For all} $A{\times}C \in store$ \\
             \qquad \begin{tabular}{l}
              \textbf{for all} $A_1{\times}C_1 \in X$      \\
                \quad \begin{tabular}{l}
                  \textbf{If} $C=C_1$ and $(A)^{(1,2,C)(1,2,C)}=(A_1)^{(1,2,C_1)(1,2,C_1)}$    \\
                   \quad $X \leftarrow X \backslash A_1{\times}C_1$,  \qquad \quad ""deleting all elements of $X$ whose implication  \\                            
                 \end{tabular}
                 \end{tabular}\textbf{\\}
              $X_1 \leftarrow X \cap X_1$,              \qquad \qquad  \qquad\qquad convey the same information as those in $store$"" \\
              $X \leftarrow X \backslash X_1$,  \\
             \textbf{For all} $A_1{\times}C_1 \in X_1$ \\
             \qquad \begin{tabular}{l}
              \textbf{for all} $A{\times}C \in X$      \\
                \quad \begin{tabular}{l}
                  \textbf{If} $C=C_1$, $A \subseteq A_1$ and $(A)^{(1,2,C)(1,2,C)}=(A_1)^{(1,2,C_1)(1,2,C_1)}$      \\
                   \quad  $X_1 \leftarrow X_1 \backslash A_1{\times}C_1$,   \qquad \qquad   ""deleting all elements of $X_1$ whose  \\ 
                                             \qquad \qquad  \qquad\qquad \quad   \qquad \qquad \qquad  implication  derive from those in $X$""
                 \end{tabular}
                 \end{tabular}\textbf{\\}
                 $X \leftarrow X \cup store \cup X_1$,  
       \item[\textbf{Return}] $X$.
 \end{description}

  We can finally propose an algorithm for the construction of an optimal base of implications $\mathcal{B}^{op}_{CAI}$ in a triadic context. This algorithm is essentially based on the construction of the set $\mathcal{UP}_{2}(\KK)$. The functions $Decomposition(..., ..., ...)$, $Simplification(...,...,...)$, $Conditional\_composition(...,...,...)$ and $Accumulation(...,...,...)$ describe exactly the simplifications (\textbf{[Decomposition], [Conditional composition], [Simplification]} and \textbf{[Accumulation]}) done in Example~\ref{exple optimalbase CAI}. Based on unit pseudo-features (the set X), the function $optimalbase_{2}(\KK, X)$ constructs the optimal base $\mathcal{B}^{op}_{CAI}$, in which each implication is considered as a triplet $(premise, conclusion, condition)$ corresponding to
  $$ premise \overset{condition}{\longrightarrow} conclusion$$

 \begin{description}
  \item[\textbf{Function} \ $optimalbase_{2}(\KK, X)$] :  
  \item[\textbf{Input}] The set $X \leftarrow mincover_{2}(\KK, Y)$ with $Y \leftarrow pseudofeat_{2}(\KK, \mathcal{N}, \mathfrak{F}(\KK))$ and the context $\KK$.   \\
    \qquad \begin{tabular}{l}
      $prem \leftarrow []$          \qquad \qquad \qquad ""$prem, cond, concl$ are tables storing respectively, \\
      $cond \leftarrow []$          \qquad \qquad \qquad   the premise, the condition and the conclusion  \\
      $concl \leftarrow []$         \qquad \qquad \qquad \qquad  of each implication"" \\
      $i \leftarrow 0$  \\
      \textbf{For all} $A{\times}C \in X$ \\
             \qquad \begin{tabular}{l}
                $prem[i] \leftarrow A$,  \\
                 $cond[i] \leftarrow C$, \\
                  $concl[i] \leftarrow A^{(1,2,C)(1,2,C)} \backslash A$,  \\
                  $i \leftarrow i+1$,
                 \end{tabular}
         \end{tabular}\textbf{\\}
         $base \leftarrow Decomposition(base)$,  \\
         $base \leftarrow Conditional\_composition(base)$,  \\
         $base \leftarrow Decomposition(base)$,  
       \item[\textbf{Return}] $base$.
 \end{description}
  
  Let us note that the algorithms for computing all pseudo-features, unit pseudo-features with respect to $\mathcal{C}$ and even our base  $\mathcal{B}^{op}_{ACI}$ are similar to what we have done. \\

The complexity of the function $feat_{2}(Z,\KK)$ depends essentially on that of the incremental construction of features (and therefore concepts) \cite{MakhalovaNourine2017}. Let us note that all concepts of the initial context $\mathbb{K}$ are known ; the only new relations in $\mathbb{K}[Z]$ are $\mathfrak{T}^{o_Z}=\{(\{o_Z\}, A_2, A_3) \in \mathfrak{T}(\{o_Z\}, M, \mathcal{C}, I^{'})$\} where $o_Z \notin G$ is the new object augmenting $\mathbb{K}$ and  $\mathfrak{T}(\{o_Z\}, M, \mathcal{C}, I^{'})$ is the set of all concepts of the sub-context  $(\{o_Z\}, M, \mathcal{C}, I^{'})$ of $\mathbb{K[Z]}$ with $I^{'}=I_{Z}\cap (\{o_Z\}{\times}M{\times}\mathcal{C})$. Thus, to build $\mathfrak{T}(\KK[Z])$ i.e. $\mathfrak{F}(\KK[Z])$, we need to merge $\mathfrak{T}(\KK)$ with $\mathfrak{T}^{o_Z}$. Thanks to \cite{MakhalovaNourine2017}, the time complexity of this process is generalized by $O((|M||\mathcal{C}|)^{2}|\mathfrak{F}(\mathbb{K})|)$. Since a feature defines a single concept, the complexity above is that of $\mathfrak{F}(\mathbb{K}[Z])$. Finally, the complexity becomes exponential as we go through $\mathcal{N}$ in the function $pseudofeat_{2}(\KK, \mathcal{N})$. We therefore need around $O(2^{|M|}(|M|^2|\mathcal{C}|^{3})|\mathfrak{F}(\mathbb{K})|)$ operations to build the set $\mathbb{P}_{i}(\KK)$ for any $i \in \{1,2\}$. This complexity is essentially that of the function $optimalbase_{2}(\KK, \mathcal{N})$.

\section{Conclusion}\label{conclusion}

In this work, which focused on constructing an optimal base for ACI and CAI, respectively, we first studied quasi-features and pseudo-features, then we proved that pseudo-features play a similar role to the pseudo-intents known in dyadic contexts. This led us to review the construction of bases for BACI and BCAI. Next, we introduce a study on unit pseudo-features. These tools helped us to build a minimal base of ACI and CAI respectively. Using the logic of simplification of implication, we then generated optimal bases for ACI and CAI. These optimal bases have a small number of implications and a high reduction rate of extraneous elements. Finally, we conclude this work by proposing a theoretical study of the complexity of these constructions. The practical study of complexity is also something we are considering programming our algorithms to offer.

\section*{Acknowledgement}

\section*{Declarations}

\subsection*{Funding}
This research received no external funding.

\subsection*{Conflict of interest/Competing interests}
The authors declare that they have no conflict of interest.

\subsection*{Ethics approval and consent to participate}
Not applicable.

\subsection*{Consent for publication}
Not applicable.

\subsection*{Data availability}
Not applicable.

\subsection*{Materials availability}
Not applicable.

\subsection*{Code availability}
Not applicable.

\subsection*{Author contributions}
Conceptualization and methodology: Leonard Kwuida; writing---original draft preparation: Romuald Kwessy Mouona; writing---review and editing: All authors.

\end{document}